\documentclass[11pt,table]{article}
\usepackage[final]{acl}
\usepackage{times}
\usepackage{latexsym}
\usepackage[T1]{fontenc}
\usepackage[utf8]{inputenc}
\usepackage{microtype}
\usepackage{colortbl}
\usepackage{amsmath, amssymb, bm}
\usepackage{graphicx}
\usepackage{algorithm}
\usepackage{algpseudocode}
\usepackage{booktabs}
\usepackage{natbib}
\usepackage{multirow}
\usepackage{subcaption}
\usepackage{soul}
\usepackage[many]{tcolorbox}
\tcbuselibrary{breakable}
\definecolor{darkTeal}{HTML}{23373b}
\definecolor{shanghai}{RGB}{252,230,213}
\definecolor{shanghai2}{RGB}{248,205,172}
\definecolor{beijing}{RGB}{218,227,245}
\definecolor{beijing2}{RGB}{182,199,234}
\definecolor{chunk}{RGB}{231,230,230}
\newcommand\dashlinefill{\leavevmode\leaders\hbox{-}\hfill\kern0pt}
\newenvironment{casestudyblock}[1]{
\tcolorbox[capture=minipage,fonttitle=\small\bfseries\color{darkTeal}, enhanced jigsaw,boxsep=1mm,colback=black!10,breakable,noparskip,
	on line,tcbox width=auto,left=0.3em,top=1mm, toptitle=0mm,
	colframe=black!20, arc=0pt,outer arc=0pt,
	opacityback=0.7,nobeforeafter,title=#1]}
{\endtcolorbox}
\newtcolorbox{prompt}{
	fontupper = \bf, 
	boxrule = 1.5pt,
	colframe = black,
	rounded corners,
	arc = 5pt   
}

\title{A Comprehensive Graph Framework \\for Question Answering with Mode-Seeking Preference Alignment}
\author{
 \textbf{Quanwei Tang\textsuperscript{1}},
 \textbf{Sophia Yat Mei Lee\textsuperscript{2}},
 \textbf{Junshuang Wu\textsuperscript{3}},
 \textbf{Dong Zhang\textsuperscript{1*}},
\\
 \textbf{Shoushan Li\textsuperscript{1}},
 \textbf{Erik Cambria\textsuperscript{4}} and
 \textbf{Guodong Zhou\textsuperscript{1}}
\\
\\
 \textsuperscript{1}School of Computer Science \& Technology, NLP Lab, Soochow University, China \\
 \textsuperscript{2}Department of Chinese and Bilingual Studies, The Hong Kong Polytechnic University \\
 \textsuperscript{3}Beijing Jinghang Research Institute of Computing and Communication, China\\
 \textsuperscript{4}College of Computing and Data Science, Nanyang Technological University, Singapore\\
 \small{
   \href{mailto: dzhang@suda.edu.cn}{ dzhang@suda.edu.cn}
 }
}
\begin{document}
\maketitle
\begin{abstract}
Recent advancements in retrieval-augmented generation (RAG) have enhanced large language models in question answering by integrating external knowledge. However, challenges persist in achieving global understanding and aligning responses with human ethical and quality preferences. To address these issues, we propose GraphMPA, a comprehensive graph-based framework with mode-seeking preference alignment. Our approach constructs a hierarchical document graph using a general similarity measurement, mimicking human cognitive processes for information understanding and synthesis. Additionally, we introduce mode-seeking preference optimization to better align model outputs with human preferences through probability-matching constraints. Extensive experiments on six datasets demonstrate the effectiveness of our  \href{https://github.com/tangquanwei/GraphMPA}{GraphMPA}.
\end{abstract}
\section{Introduction}
\renewcommand{\thefootnote}{*} 
\footnotetext{Corresponding Author: Dong Zhang}

Retrieval-augmented generation (RAG) with large language models (LLMs) has recently emerged as a promising approach in question-answering (QA)
\cite{zhao2024retrievalaugmentedgenerationaigeneratedcontent,gaoRetrievalAugmentedGenerationLarge2024}. This is mainly due to its ability to retrieve external documents, thus increasing the knowledge of the model. 
However, despite its advances, existing studies still face the following challenges at both input and output levels.
\begin{figure}[t]
    \centering
    \includegraphics[width=1\linewidth]{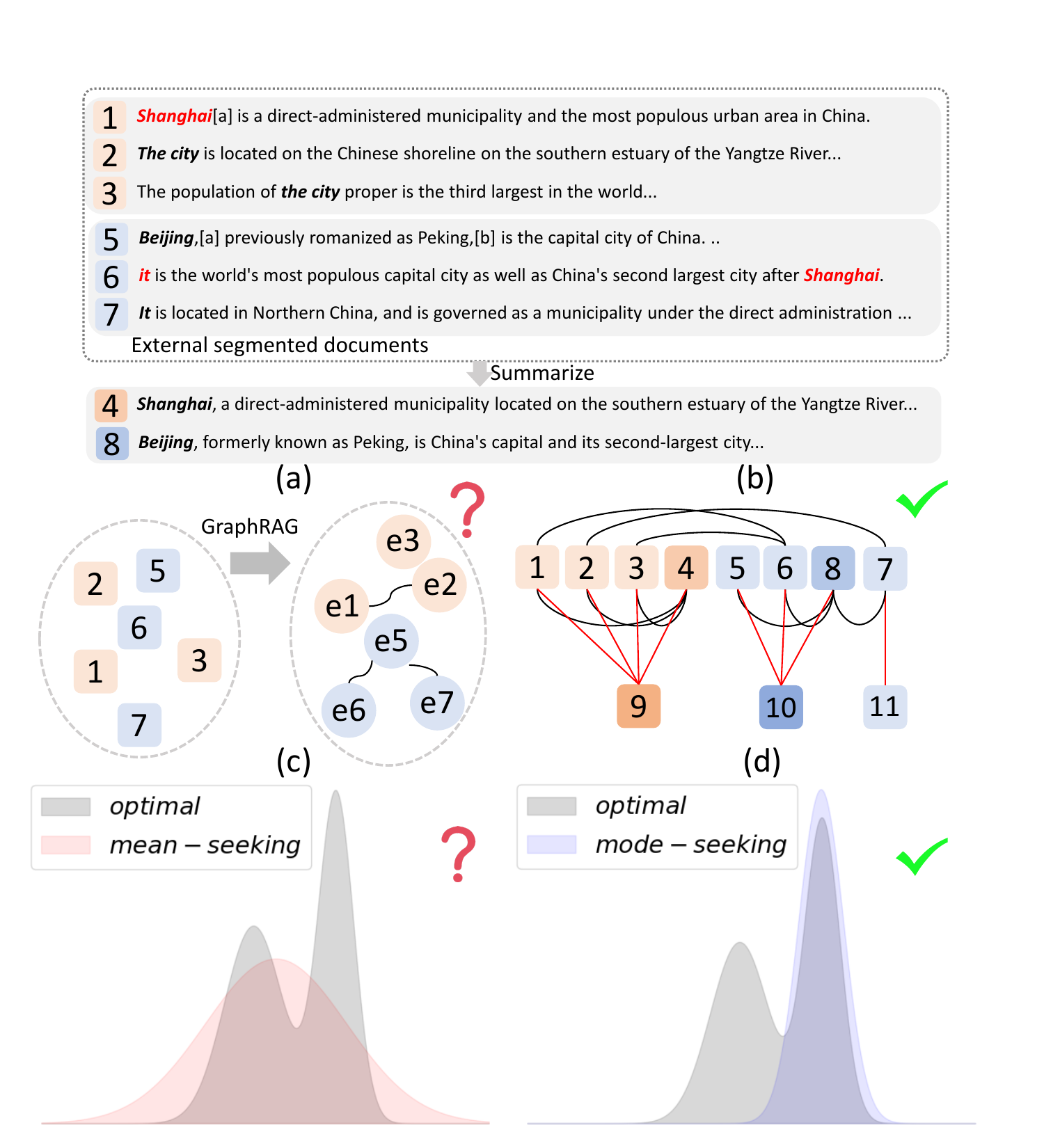}
    \caption{(a) Prior entity-based graph and (b) Our hierarchical graph with community summarization. (c) Prior DPO with large-scale LLMs generated data and (d) Our MS with small-scale LLMs synthetic data.}
    \label{fig:intro}
\end{figure}
From the representation and understanding of \textbf{input external documents}, previous works 
\cite{barnett2024sevenfailurepointsengineering,asaiSELFRAGLEARNINGRETRIEVE2024,
xuUnsupervisedInformationRefinement2024,
lewisRetrievalAugmentedGenerationKnowledgeIntensive2021,liudis}
often struggle with tasks that require comprehensive, global understanding, as they fail to provide a unified and holistic view of the relevant external knowledge. For instance, as shown in Figure~\ref{fig:intro}, if we directly retrieve effective information across all input documents for an overarching query like "What about Shanghai?", the potential answer may refer to sentence \sethlcolor{beijing} \hl{6}, where the pronoun "it" is interpreted as referring to "Shanghai" (it actually refers to Beijing.) due to the presence of the query words at the end of this sentence. Only very recently,~\citet{guo_lightrag_2024} and~\citet{edgeLocalGlobalGraph2024} introduced a global graph-based RAG strategy to alleviate this issue, but it heavily relies on large-scale LLMs (e.g., GPT-4) to extract entity pairs and their corresponding relations from each sentence as Figure~\ref{fig:intro}(a). This not only consumes significant resources but may also overlook sentences without entities or relations like the sentence \sethlcolor{shanghai} \hl{3} in Figure~\ref{fig:intro}. Therefore, we argue for the use of smaller, more efficient LLMs and the construction of sentence relevance using a more general and comprehensive measurement as Figure~\ref{fig:intro}(b).
From the quality of \textbf{output answers}, ensuring that LLMs generate responses aligned with human standards and preferences remains a significant statistical and practical challenge in QA~\cite{onesta,camsta}, as it involves modeling complex patterns of human reasoning, contextual understanding, and nuanced language use.
Although approaches like DPO~\cite{rafailov2024directpreferenceoptimizationlanguage} have been proposed to mitigate this issue, they lead to a compromised approximation of optimal distribution~\cite{naacl/LinJLGE21, acl/Wu0DSMH24} based on large-scale LLMs (GPT-4) generated corpus, such as a \textit{mean-seeking} policy that places large mass to the mean of different modes~\cite{jmlr/0001SLKMW22} as illustrated in Figure~\ref{fig:intro}(c). 
However, it is insufficient since we need to estimate the exact and primary mode of the target distribution~\cite{ji_towards_2024}. We argue that a \textit{mode-seeking} strategy based on an auto-constructed corpus with small-scale LLMs, can more effectively capture the alignment with human preferences as shown in Figure~\ref{fig:intro}(d).

To address these challenges, in this paper, we propose a comprehensive hierarchical graph 
framework with mode-seeking preference alignment (\textbf{GraphMPA}) based on RAG for QA. Specifically, we first demolish external knowledge into a hierarchical graph using a simple but comprehensive similarity measurement, mimicking human cognitive processes in organizing and synthesizing information through abstractive summarization from low-level to high-level~\cite{SWELLER1988257,chandler1991cognitive}. Next, we retrieve the top-K small documents to construct a human-preferred dataset by data synthesis for real preference alignment. Finally, to achieve better mode-seeking, we introduce mode-seeking preference optimization to align the model with human preferences. In general, our contributions are summarized as follows:

1) We propose a hierarchical graph with community summarization based on a general similarity measurement, improving comprehensive sentence relevance and global understanding of external candidate knowledge for QA tasks.

2) We introduce a mode-seeking preference optimization strategy by applying probability-matching constraints between the parametrized policy and the optimal policy without relying on very large-scale LLMs to generate human-preferred data.

3) We carry out extensive experiments and detailed analysis on six representative datasets, demonstrating the effectiveness of our proposed GraphMPA.

\section{Related Work}

LLMs have achieved significant success in a wide range of QA tasks.
However, their notable challenge is the limited access to specialized or up-to-date knowledge, which can lead to outdated or incomplete responses in domain-specific tasks. To address these issues, RAG has been proposed as a solution~\cite{gaoRetrievalAugmentedGenerationLarge2024}.  
Our work is mainly related to RAG and human preference alignment.

\textbf{Retrieval-Augmented Generation.} 
RAG enhances LLMs with external retrieval, making it ideal for knowledge-intensive tasks~\cite{huafun}.
Traditional approaches of QA~\cite{tang2024questqueryawaresparsityefficient} normally face the "lost in the middle" problem~\cite{liu2023lostmiddlelanguagemodels}, and are expensive and slow. RAG can alleviate these issues by retrieving relevant documents from external sources and incorporating this information into the generation process.
However, determining when and which documents to retrieve and global understanding issues remain challenging~\cite{gaoRetrievalAugmentedGenerationLarge2024}.
Therefore,~\citet{edgeLocalGlobalGraph2024} propose a graph-based method for query-focused summarization by extracting an entity knowledge graph with very large-scale LLMs, like GPT-4. Meanwhile,~\citet{sarthiRAPTORRecursiveAbstractive2024} suggest a recursive, hierarchical summarization approach by a tree structure, but it ignores the internal relations among nodes in each layer.
Unlike the above studies, we design a comprehensive graph structure with simple and fine-grained document relations to incorporate effective knowledge, which completely abandons large-scale LLMs as entity-relation extractors. 

\textbf{Human Preference Alignment (HPA).}
Recently, Direct Preference Optimization (DPO)~\cite{rafailov2024directpreferenceoptimizationlanguage} has emerged as a promising alternative to the previously popular reinforcement learning from human feedback (RLHF)~\cite{christiano2023deepreinforcementlearninghuman, ouyang2022traininglanguagemodelsfollow}. 
It simplifies the training pipeline, eliminating the need for separate reward models and policy optimization ~\cite{rafailov2024directpreferenceoptimizationlanguage}.
However, this often results in a mean-seeking approximation that compromises the optimal solution and relies on GPT-4 generated data for optimization. Consequently, the strategy $\pi_{\theta}$ tends to cover all modes of the target strategy $\pi^{\ast}$, rather than concentrating on the most important modes. As a result, the generated texts or behaviors may lack clear direction or focus, failing to effectively capture the key features of human preferences.
Unlike previous works, 
to avoid the compromised mean-seeking approximation of the optimal solution in standard DPO, we introduce the mode-seeking (MS) loss inspired by~\cite{ji_towards_2024} and design an intuitive auto-constructed dataset with small-scale LLMs for training.

\section{Methodology}

\textbf{Motivation.}
To systematically understand the traditional fragmented pieces of knowledge, we draw inspiration from human cognitive processes in handling and organizing information, proposing to summarize closed small documents and organize them into \emph{a hierarchical graph}. This also enhances both the low-level and the high-level comprehension of external knowledge. Then, we simulate the human thinking process to derive the final answer through structured reasoning. This method leverages the \emph{mode-seeking preference optimization} aiming to better fit the optimal pattern (mode) rather than the overall expectation.

\textbf{Task Formulation.}
In RAG, the objective is to answer a natural language question $Q$ by querying a set of segmented documents $D$. Each document $d$ is typically derived from external text. The question and the retrieved documents are then passed into an LLM $\mathcal{M}$ to generate the answer $A$. During the retrieval process, an embedding model $\textsc{Embed}$ is used to transform the text input into dense vector representations. 
\subsection{Summarization-based Hierarchical Graph}
\begin{figure}[t]
    \centering \includegraphics[width=1\linewidth]{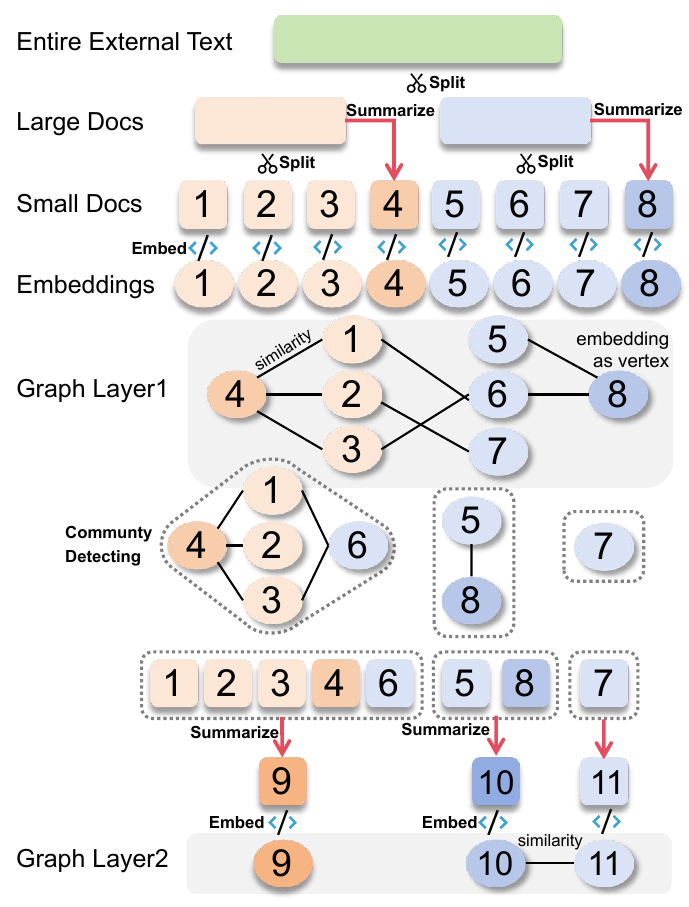}
    \caption{The process of building the hierarchical graph with community summarization.
    }
    \label{fig:build}
\end{figure}

We expect to capture both low-level and high-level information in the input ultra-long text, so we intersperse summarization strategies to construct a hierarchical graph network. This summarization allows retrieval augmentation of context at different scales. Figure~\ref{fig:build} shows the overall iterative process of building a graph, including graph initialization, community detection, and hierarchical structuring, which collectively enhance the organization and representation of retrieved information.

\textbf{Document Splitting and Relations.} 
The entire input candidate text to be retrieved is first divided into large documents (e.g., the whole of sentences 1, 2 and 3 in Figure~\ref{fig:intro}), each of which is then summarized to produce a summary document denoted as $S_D = \{D_1, \cdots, D_l\}$ with LLMs. These large documents are recursively subdivided into smaller documents $S'_D = \{D_k, \cdots, D_n\}$ (e.g., the sentence 1 or 2 in Figure~\ref{fig:intro}) with $\text{TextSplit}(D_{large},small)$. For each small document $D_i$, the document embedding $e_{D_i}$ is computed using an embedding model \textsc{Embed}, like BGE-M3 ~~\cite{chen2024bgem3embeddingmultilingualmultifunctionality,alwara}, which transforms the textual content of the document into a dense vector representation. The similarity between the embeddings of documents is then computed by cosine similarity as,
\begin{equation}
\text{sim}(D_i, D_j) = \frac{e_{D_i} \cdot e_{D_j}}{\|e_{D_i}\| \|e_{D_j}\|}
\end{equation}
To incorporate all proper information, we combine all summary documents $S_D$ and small documents $S'_D$ as $\bf D$=$\{D_1,\cdots, D_n\}$. Then, a document similarity matrix $ M $ is constructed by calculating the pairwise relations among all documents' embeddings.

\textbf{A Layer of Graph Building.}  
For the convenience of graph calculation, we remove a small number of edges with weak correlation by a threshold $\tau$ according to the similarity matrix $ M $.
To this end, we construct a layer of the graph, defined as $\mathcal{G} = (\mathcal{V}, \mathcal{E})$:

\textbf{Vertexes $\mathcal{V}$:} $v\in \mathcal{V}$ represents a small document or summarization document $D$ from large document with its embedding as input. Formally,
\begin{equation}
    v = e_D = \textsc{Embed}(D)
\end{equation}

\textbf{Edges $\mathcal{E}$:} $e\in \mathcal{E}$ represents a similarity between two small documents through embeddings. 
The weight of edge $e$ is defined:
\begin{equation}
w_{ij}=\left\{\begin{matrix}
    s_{ij}& ,&\text{sim}(D_i, D_j) \ge  \tau\\
    0 &,& else
\end{matrix}\right.
\end{equation}
This layer of the graph is then added to the set of $ layers $. 

\textbf{Community Detection.}
To better understand complex and lengthy candidate text, we construct a hierarchical graph. We cluster nodes in the current layer to organize segments into cohesive groups, referred to as community detection (CD). These groups are then summarized into a new node for the next layer.
To achieve this, we apply the Leiden algorithm~\cite{traag2019louvain} to each graph layer, extracting communities. The algorithm ensures internal connectivity and finds high-quality partitions efficiently, making it ideal for large-scale, complex networks. This step clusters related contexts, aiding the retrieval process. Formally,
\begin{equation}
C = \text{CommDetect}(G_i)
\end{equation}
For each detected community, a community summary $ S_C $ is generated, forming a new document $ D' $. This document encapsulates the key aspects of the community’s content, and a new graph $G_{i+1}$ is created based on these summaries. The graph is added to the set of $ layers $ layers, and the embedding calculation for each graph is repeated until the desired depth $ L $ is reached.

\textbf{Iterative Hierarchy Building.}
At each iteration, document embeddings are recalculated, and document similarities are updated. This recursive process refines the system's understanding of the document space, improving retrieval and performance in tasks like query answering. The whole process can refer to Algorithm~\ref{build}. The argument $ large $ denotes the length of large chunks (an integer), $ small $ represents small documents length (an integer), $LLM.summary$ denotes the abstractive summarization by LLMs, and $ k $ indicates the number of top $ k $ nodes retrieved for related node searching. 
\begin{algorithm}[t]
\caption{Build Graph Algorithm}
\label{build}
\begin{algorithmic}[1]
\Function{Build Graph}{
$text$ \Comment{Document}, 
$large, small$,  \Comment{Output length}
$n\_layers$, \Comment{Depth $ L $}
$\tau$ \Comment{Threshold}
}
\State $D_{large} \gets \text{TextSplit}(text,large)$
\State $S_D \gets \text{LLM.summay}(D_{large},small)$
\State $S_D'\gets \text{TextSplit}(D_{large},small)$
\State $D \gets S_D \cup S_D'$
\State $layers \gets []$
\While{$n\_layers >0$}
\State $e_D \gets \textsc{Embed}(D)$
\State $M \gets \text{sim}(e_D,e_D^\text{T})$
\State $sims \gets \text{sort}(M, axis \gets 0, reverse)$
\State $\mathcal{V} \gets \text{arg~sort}(M, axis \gets 0, reverse)$
\State $\mathcal{E} \gets []$
\For{$u \gets 0$ to $\text{len}(\mathcal{V})$}
    \For{$v ~~\text{in} ~~ \mathcal{V}[u]$}
        \State $w \gets sims[u][v]$
        \If {$w \ge \tau$}
        \State $\mathcal{E}.append((u, v, w))$
        \EndIf
    \EndFor
\EndFor
\State $\mathcal{G} \gets (\mathcal{V},\mathcal{E})$
\State$layers.\text{append}(\mathcal{G})$
\State $n\_layers \gets n\_layers-1$
\Comment{Prepare for the next layer}
\State $C \gets \text{CommDetect}(\mathcal{G})$ 
\State $D \gets \text{LLM.summay}(C,small)$
\EndWhile
\State \Return{layers} 
\EndFunction
\end{algorithmic}
\end{algorithm}

\subsection{Generating with Preference Alignment}

In this section, we introduce a preference alignment approach with a novel optimization strategy that simultaneously emphasizes intermediate reasoning steps and the final answer, enhancing the model’s understanding of the reasoning process and enabling more reliable, coherent, and context-aware outputs.
Specifically, we first retrieve the most query-relevant small documents. Then, based on these informative documents, we build the contrastive dataset with the \textbf{human-preferred answer} with a reasoning process and the \textbf{less preferred answer} without any reasoning process. Finally, we propose to leverage the mode-seeking loss to optimize the model more satisfying against traditional DPO.

\textbf{Retrieval with Semantic Ranking.} 
After understanding both low- and high-level messages of the input long candidate text, we select the most relevant information for the specific query by retrieving top-k small documents on semantic measurement.
To this end, we first calculate the query embedding for subsequent small document matching and ranking:
\begin{equation}
e_q = \textsc{Embed}(q)
\end{equation}
Next, we calculate the similarity between the query embedding and all documents at each layer. The top $ k $ most similar documents are then selected: 
\begin{equation}
\text{results} = \max_{D_{\text{layer}} \in layers } \, \text{sim}(e_q, D_{\text{layer}})
\end{equation}
where the query embedding $ e_q $ is compared with each document $ D_{\text{layer}} $ across all layers. The similarity function $ \text{sim} $ measures the semantic closeness between the query and the document at each layer. The $ k $ documents with the highest similarity scores are selected as the most relevant results. This method enables more efficient retrieval by leveraging the multi-layer structure of the documents.
\begin{figure}
    \centering
    \includegraphics[width=1\linewidth]{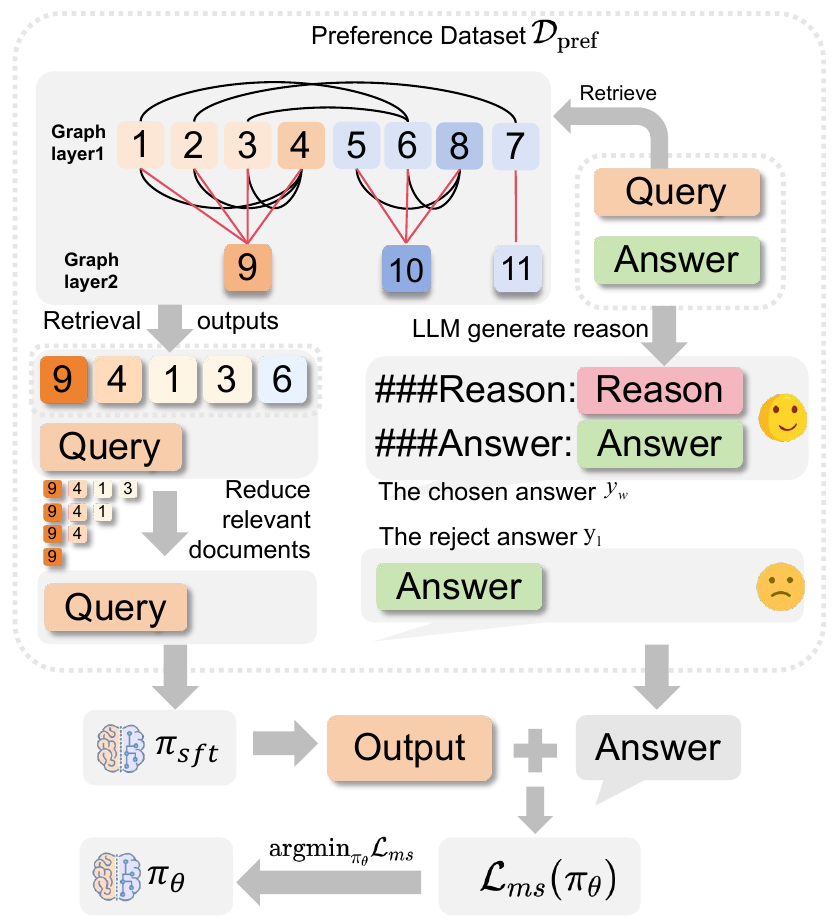}
    \caption{The human preference alignment training process with MS loss.}
    \label{fig:dpo}
\end{figure}

\textbf{Preferred Dataset Preparation.}
First, unlike previous works~\cite{wang-etal-2024-improving-text_gpt_gen_data} use GPT ~\cite{openai2024gpt4technicalreport} to generate train data, we yield reasoning explanations using the query and answer with multiple small-scale LLMs (i.e. Qwen2.5-7B ~\cite{qwen2025qwen25technicalreport}, LLaMA3-8B~\cite{grattafiori2024llama3herdmodels} and Mistral-8B~\cite{ministraux_update} with Chain of Thought (CoT) ~\cite{wei2023chainofthoughtpromptingelicitsreasoning}. Note that our method of synthesizing data consumes very few resources, compared with the previous direct use of large-scale GPT-4 to generate data in DPO.
Second, for each query $ q $, we expect it to have different types and relevant contexts $ C $ to construct more diverse samples for learning.
Therefore, we use $k\in [0, \text{len}(C)]$ value to select contexts ranging from strongly correlated to weakly correlated: $C_1$ = [doc$_9$], $C_2$ = [doc$_9$,doc$_4$], $C_3$ = [doc$_9$,doc$_4$,doc$_1$] $\cdots$ as illustrated in Figure~\ref{fig:dpo}. 
We define two types of outputs: the positive item $ y_w $, which includes both the reasoning process and final answer, and the negative item $ y_l $, which only includes the answer in the training set. The overall data format is as follows:
\begin{equation}
\mathcal{D}_{\text{pref}} = \{ (q^{(i)}, C^{(i)}, y_w^{(i)}, y_l^{(i)}) \mid i \in \{1, 2, \cdots, N\} \}
\end{equation}
Where $N$ denotes the number of queries in the training set. An example of train data is in Appendix~\ref{train_data}.

\textbf{Preference Alignment Training.}
With the above two types of data, we hope that our model is more inclined to generate answers that are reasonable and well-founded, that is, answers that are human-preferred.
Building on the success of Direct Preference Optimization (DPO)~\cite{rafailov2024directpreferenceoptimizationlanguage} with reinforcement learning in the training of LLMs, we propose to seek the mode of real distribution with the mode-seeking loss function (MS), rather than mean-seeking in DPO~\cite{jmlr/0001SLKMW22,ji_towards_2024}. It is defined as:

\begin{equation}
\begin{split}
\mathcal{L}_{\text{ms}}(\pi_\theta) &= \mathbb{E}_{x \sim \mathcal{D}_{\text{pref}}} \mathbb{E}_{\pi_{\text{sft}}}(\mathbf{y}_{1:K} | x) \Big[ \\
& \mathbb{D}_{\text{KL}}(p_{f_\theta}(\cdot | \mathbf{y}_{1:K}, x) \| p_{r_\phi}(\cdot | \mathbf{y}_{1:K}, x)) \Big]
\end{split}
\end{equation}

where $\mathcal{D}_{\text{pref}}$ denotes the preference dataset which contains human-labeled preference data. 
$\pi_{\text{sft}}(y_{1:K}|x)$ indicates the supervised fine-tuned (SFT) policy and $\pi_{\text{sft}}$ means the probability distribution of responses given a prompt $x$ after the language model has undergone supervised fine-tuning.
$p_{f_\theta}(\cdot | y_{1:K}, x)$ represents the empirical distribution based on the model policy. 
$p_{r_\phi}(\cdot | y_{1:K}, x)$is the empirical distribution based on the reward model. 

\begin{table*}[t!]
\small
\centering
\begin{tabular}{l|llllll}
\toprule
& QASPER          & QuALITY        & RiddleSense         & PubMedQA          & MedQA          & MedMcQA          \\ \hline
\multicolumn{7}{c}{RAPTOR~\cite{sarthiRAPTORRecursiveAbstractive2024}}      \\ \hline
{\textbf{LLaMa 8B}}               & 0.3657              & 45.62          & 49.66          & 58.40          & 53.10          & 50.84          \\ \hline
\multicolumn{7}{c}{LightGraphRAG~\cite{guo_lightrag_2024}}      \\ \hline
\textbf{LLaMa 8B}                   & 0.3585          & 45.82          & 50.83          & 49.00          & 45.18          & 50.91          \\ \hline
\multicolumn{7}{c}{Reward-RAG~\cite{nguyenRewardRAGEnhancingRAG2024}}      \\ \hline
\textbf{GPT-3.5-turbo}                      & -               & -              & -              & 69.20          & 59.20          & 52.40          \\
\textbf{GPT-4-0613 }                      & -               & -              & -              & 70.80          & 64.50          & 57.40          \\ \hline
\multicolumn{7}{c}{LLaMa 8B~\cite{grattafiori2024llama3herdmodels}}      \\ \hline
\textbf{Basic LLM}                             & 0.1040          & 32.10          & 62.78          & 49.60          & 60.17          & 50.01          \\
\textbf{Basic RAG}                        & 0.3599          & 41.73          & 60.24          & 68.80          & 57.34          & 50.35          \\
\rowcolor{orange!10}\textbf{GraphMPA (ours) }                     & \textbf{0.3775} & \textbf{47.05} & \textbf{73.65} & \textbf{73.00} & \textbf{66.54} & \textbf{64.28} \\ \hline
\multicolumn{7}{c}{Qwen 7B~\cite{qwen2025qwen25technicalreport}}      \\ \hline
\textbf{Basic LLM}                              & 0.088           & 41.54          & 65.95          & 28.60          & 52.00          & 53.36          \\
\textbf{Basic RAG}                         & 0.2654          & 47.32          & 65.95          & 50.67          & 50.82          & 55.15          \\
\rowcolor{orange!21}\textbf{GraphMPA (ours) }                               & \textbf{0.3734}   & \textbf{47.64} & \textbf{71.79}  & \textbf{71.92} & 61.96 & \textbf{57.61} \\ \hline
\multicolumn{7}{c}{Mistral 8B~\cite{ministraux_update}}      \\ \hline
\textbf{Basic LLM}                           & 0.1135          & 35.32          & 59.39          & 47.40          & 54.28          & 58.07          \\
\textbf{Basic RAG}                      & 0.3228          & 43.64          & 64.29          & 66.80          & 56.64          & 63.93          \\
\rowcolor{orange!32}\textbf{GraphMPA (ours) }                   & \textbf{0.3873}   & \textbf{51.76} & \textbf{73.92}      & \textbf{72.82}      & \textbf{68.66}      & \textbf{67.06}     \\
\bottomrule
\end{tabular}
\caption{
 Performance comparison of various models and approaches across different QA datasets. The marker '-' denotes the results unavailable in public reports.
}
\label{tab:exp_all}
\end{table*}

\begin{table}[t]
\centering
\resizebox{\linewidth}{!}{
\begin{tabular}{lllll} \toprule
Dataset & Docs  & TrainQAs & TestQAs & Avg Tokens \\ \midrule
QASPER  & 6724  & 5000     & 3328    & 358        \\
QuALITY & 5772  & 4600     & 2128    & 365        \\
PubMed  & 1500  & 5000     & 500     & 398        \\
Riddle  & 3510  & 5000     & 1021    & 150        \\
MedQA   & 3765  & 5000     & 1273    & 194        \\
MedMC   & 16777 & 5000     & 4183    & 175        \\ \bottomrule
\end{tabular}
}
\caption{The dataset statistics include the number of small documents, train QA pairs, test QA pairs, and average document tokens. }
\label{tab:dataset_stats}
\end{table}

\section{Experiments}
We conduct experiments to evaluate our method on diverse QA tasks against baselines.
\subsection{Experimental Setting}

\textbf{Datasets.} We evaluate various QA datasets, roughly divided into: 
1) \textbf{GenerativeQA}: QASPER~\cite{dasigi2021datasetinformationseekingquestionsanswers}, evaluated with the \textit{ROUGE} score (Recall-Oriented Understudy for Gisting Evaluation)\cite{lin-2004-rouge,ganesan2018rouge20updatedimproved}; 
2) \textbf{ChoiceQA}: QuALITY~\cite{pang2022qualityquestionansweringlong} and RiddleSense~\cite{lin2021riddlesensereasoningriddlequestions}, evaluated with standard \textit{Accuracy}; 
3) \textbf{MedicalQA}: PubMedQA~\cite{jin2019pubmedqadatasetbiomedicalresearch}, MedQA~\cite{jin2020diseasedoespatienthaveMedQA} and MedMCQA~\cite{pal2022medmcqalargescalemultisubject}, evaluated with \textit{MIRAGE} (Medical Information Retrieval-Augmented Generation Evaluation)~\cite{xiong2024benchmarkingretrievalaugmentedgenerationmedicineMirage}
\footnote{\url{https://github.com/mirage-project/mirage}}. The statistical summary can refer to Table~\ref{tab:dataset_stats}, where TrainQAs represents the original split of the training set in each dataset. We use TrainQAs to automatically synthesize 20000 samples for all datasets to conduct subsequent preference alignment training.

\textbf{Baselines and Implementation Details.} 
We compare the following representative SOTA on RAG and QA: Standard RAG~\cite{lewisRetrievalAugmentedGenerationKnowledgeIntensive2021,
gaoRetrievalAugmentedGenerationLarge2024}, RAPTOR~\cite{sarthiRAPTORRecursiveAbstractive2024}, LightGraphRAG~\cite{guo_lightrag_2024} and Reward-RAG~\cite{nguyenRewardRAGEnhancingRAG2024}. In our approach, we use three backbones: LLaMa-3.1-8B-Instruct~\cite{grattafiori2024llama3herdmodels, patterson2022carbonfootprintmachinelearning}, (i.e., LLaMa 8B), Qwen2.5-7B-Instruct~\cite{qwen2025qwen25technicalreport} (i.e., Qwen 7B), and Mistral-8B-Instruct-2410~\cite{ministraux_update} (i.e., Mistral 8B). We implement training using the trl library, configuring the learning rate to 1e-5, the batch size to 4, and training for 2 epochs.
\subsection{Main Results}
\begin{table*}[t]
\small
\centering
\begin{tabular}{lccccccc}
\toprule
            & QASPER                      & QuALITY                    & RiddleSense                 & PubMedQA                  & MedQA                      & MedMcQA                     \\ \midrule
w/o Summarization   & 0.3599                      & 41.73                      & 60.24                       & 68.80                     & 57.34                      & 50.35                       \\
            & $\downarrow$ \small{0.0176} & $\downarrow$ \small{ 5.32} & $\downarrow$ \small{13.41}  & $\downarrow$ \small{ 4.2} & $\downarrow$ \small{9.2 }  & $\downarrow$ \small{ 13.93} \\ \hline
w/o Retrieval & 0.1040                      & 32.10                      & 62.78                       & 49.60                     & 60.17                      & 50.01                       \\
            & $\downarrow$ \small{0.2735} & $\downarrow$ \small{14.95} & $\downarrow$ \small{10.87 } & $\downarrow$ \small{23.4} & $\downarrow$ \small{6.37 } & $\downarrow$ \small{14.27 } \\ \hline
w/o Training  & 0.3694                      & 46.65                      & 71.79                       & 71.40                     & 63.47                      & 64.16                       \\
            & $\downarrow$ \small{0.0081} & $\downarrow$ \small{ 0.4 } & $\downarrow$ \small{1.86}   & $\downarrow$ \small{ 1.6} & $\downarrow$ \small{3.07 } & $\downarrow$ \small{0.12 }  \\ \hline
w/ DPO &0.3599	&46.06	&73.20	&71.60	&64.28	&64.16 \\ 
 & $\downarrow$ \small{0.0176} & $\downarrow$ \small{ 0.99} & $\downarrow$ \small{0.45}   & $\downarrow$ \small{  1.4} & $\downarrow$ \small{2.26 } & $\downarrow$ \small{0.12 }  \\ \hline
\rowcolor{orange!10} GraphMPA LLaMa 8B   & \bf 0.3775                      & \bf 47.05                      & \bf 73.65                       & \bf 73.00                     & \bf 66.54                      & \bf 64.28                       \\ \bottomrule
\end{tabular}
\caption{The ablation study results examine the impact of removing the retrieval, summary, and training components from our model. Performance is evaluated across six datasets: QASPER, QuALITY, RiddleSense, PubMedQA, MedQA and MedMcQA. The results show that removing the retrieval component causes the largest drop in performance, followed by the removal of the summary and training components. The full model (ours) achieves the highest performance on all datasets, highlighting the importance of each component in enhancing model effectiveness.}
\label{ablation}
\end{table*}
The experimental results presented in Table~\ref{tab:exp_all} provide a thorough comparison of our proposed GraphMPA with several representative SOTA across various datasets. From this table, we can find that:

\textbf{It is necessary to design our GraphMPA.} 
With the same backbone LLaMa 8B, two competitive methods RAPTOR and LightGraphRAG are inferior to our GraphMPA. This is mainly because although RAPTOR models external knowledge in a hierarchical tree, it ignores the associations among small documents in each layer (sibling and cousin nodes). Meanwhile, although LightGraphRAG adopts a graph structure based on extracted entities and their relationships as~\citet{edgeLocalGlobalGraph2024}, it overly relies on the ability to extract entities and relationships and may lose document information without entities. Additionally, Reward-RAG with very large-scale LLMs (GPT-3.5 and GPT-4), performs better than RAPTOR and LightRAG though, it still indicates worse performance than our approach. This may be due to outdated human preference optimization strategies like RLHF. This suggests that we should design a comprehensive graph framework and employ an advanced human preference alignment mechanism.

\textbf{RAG-based technique is worth further exploration as our GraphMPA.} 
Compared to the basic LLMs with different backbones, the addition of RAG technology improves performance. This indicates that RAG is indeed effective for QA. Besides, as our GraphMPA with improved the RAG-based technique, the performance has further improved. This suggests that RAG-based technology is worth further exploration and improvement.

\textbf{Our GraphMPA performs robustly in different frameworks.}
We evaluate GraphMPA using three different LLMs: LLaMa 8B, Qwen 7B, and Mistral 8B. Among these, the 8B LLMs significantly outperform prior models, while Qwen 7B surpasses all other models, excluding GPT-4. This improvement is largely attributed to the increased parameter size of the 8B models, which leads to better performance. These results indicate that GraphMPA exhibits strong robustness across different model architectures.
\subsection{Analysis and Discussion}

\textbf{Ablation Study.}
Table~\ref{ablation} displays the results of removing the key components in our approach. From this table, we can see that removing any module will result in significant performance degradation. This indicates that every module designed in our method is important. Among them, the removal of the summarization and retrieval modules results in the most severe performance degradation. This indicates that it is crucial to effectively represent external knowledge and retrieve relevant knowledge. Therefore, this paper designs a comprehensive hierarchical graph framework to better accomplish representation and retrieval.

\textbf{Impact of $Layers$ Scale in Graph.}
\begin{figure}[t]
    \centering
    \includegraphics[width=1\linewidth]{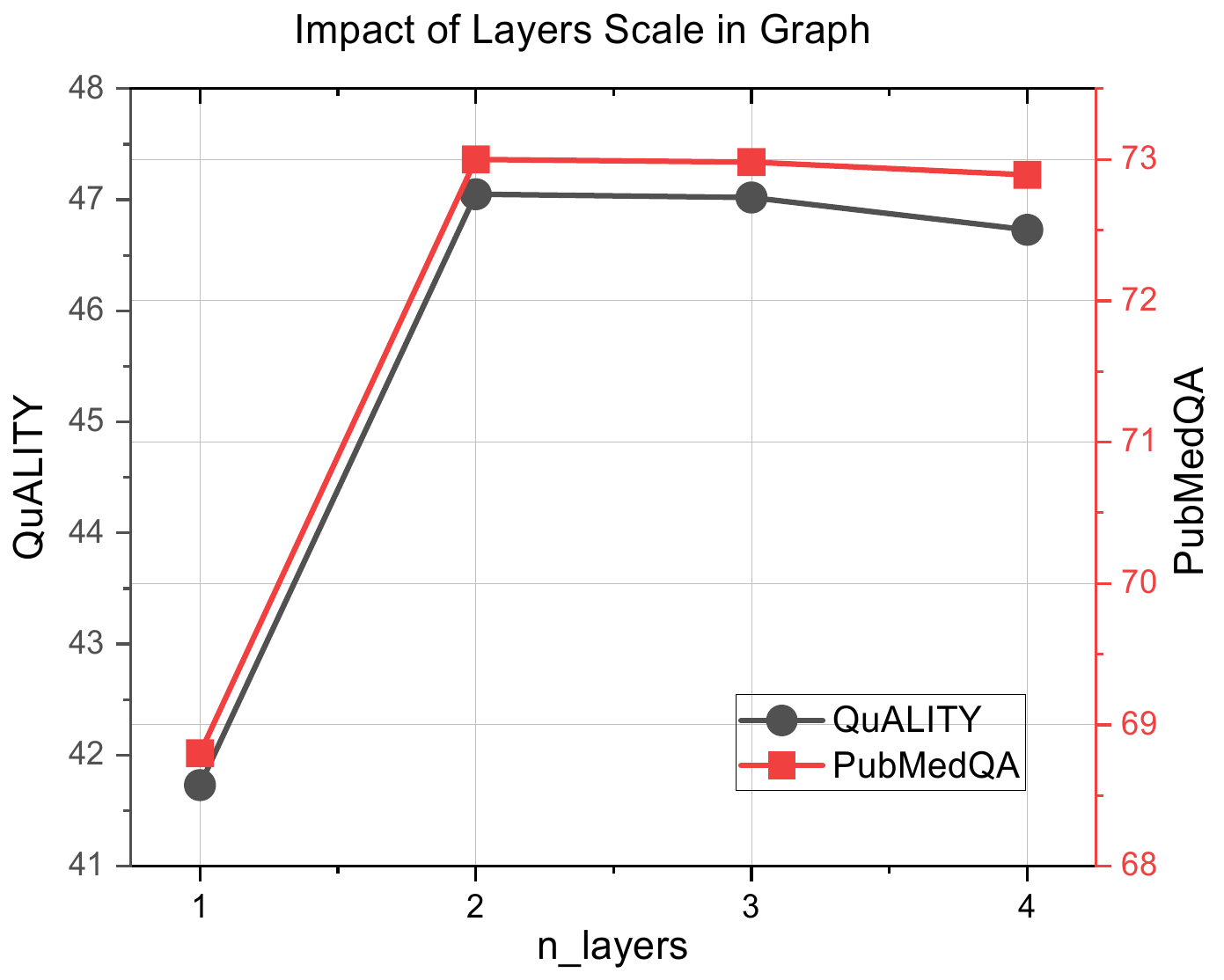}
     \caption{Performance comparison of our GraphMPA on both QuALITY and PubMedQA, as graph layers $Layers$ changes. }
     \label{fig:n_layers}
\end{figure}
Figure~\ref{fig:n_layers} displays the performance of our GraphMPA as graph layers increasing, where $n\_layers = 1$ functions as a basic RAG. From this figure, we can observe that the performance on each dataset first improves at layer 2 and then tends to stabilize or even slightly decrease. This suggests the effectiveness of our designed hierarchical graph with community summarization. However, excessive summarization when building a deep graph does not always lead to better results since too many iterations of summarization may render the document abstract and less meaningful.

\begin{figure}[t]
    \centering
    \includegraphics[width=1\linewidth]{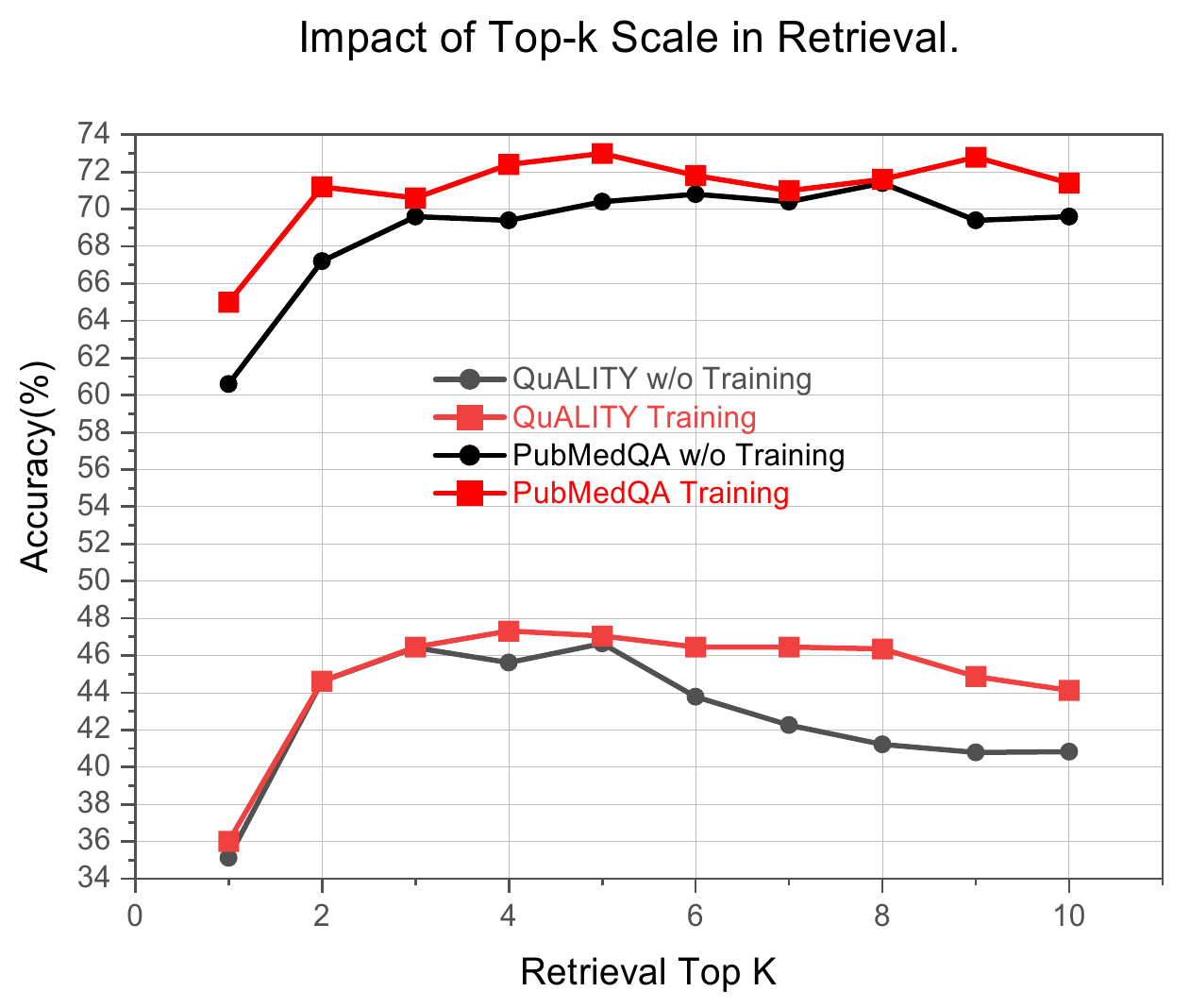}
    \caption{Performance comparison on both trained and untrained models with varying values of retrieval top-k regarding both QuALITY and PubMedQA datasets. 
    }
    \label{fig:case_topk}
\end{figure}

\textbf{Impact of Top-$k$ Scale in Retrieval.} 
In Figure~\ref{fig:case_topk}, as the number of retrieved top $k$ increases, the performance of different models generally improves first and then decreases on each dataset. This is mainly because the number of unrelated contexts increases when $k$ increases, which affects the uncertainty of the model's generation. Additionally, we observe that the accuracy of the untrained model decreases fast, while the trained model decreases slowly. This indicates that the training allows the model to extract relevant information and discriminate both related and unrelated contexts for better answers. 
Moreover, statistical analysis of the distribution of top-k documents across different graph layers is available in Appendix~\ref{sa_topk}. And the importance of ranking in Appendix~\ref{iranking}.

\begin{figure}
    \centering
    \includegraphics[width=1\linewidth]{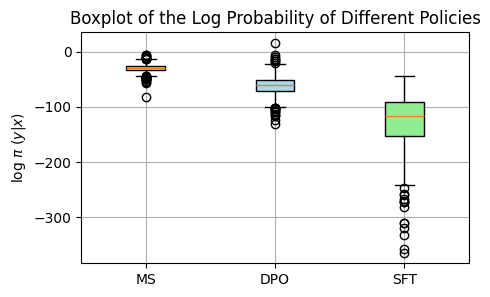}
    \caption{The log probabilities on PubMedQA with the trained model using MS, DPO and non-trained SFT model. }
    \label{fig:box_plot}
\end{figure}

\textbf{Effectiveness of mode-seeking in generation.}
The log probability $log ~\pi~ (y | x)$ is used in alignment tasks to compute the KL divergence or other optimization objectives, ensuring that the optimized policy $\pi~ (y|x)$ remains consistent with human preferences~\cite{jmlr/0001SLKMW22,ji_towards_2024}. In other words, the higher the value, the more in line with human preferences. Therefore, for each sample in PubMedQA, we calculate the log probabilities of $\pi_{ms}(y|x)$, $\pi_{dpo}(y|x)$ and  $\pi_{sft}(y|x)$, as illustrated in Figure~\ref{fig:box_plot}. From this figure, we can see that the median of our MS strategy outperforms the other two, suggesting a better human preference alignment. From the perspective of sample distribution, we can observe that each data point of our MS loss is concentrated in a small range to seek the main mode, rather than being discrete like DPO data points, striving for a comprehensive (mean) mode. Meanwhile, SFT discretizes to more remote areas and cannot even achieve the average mode.

\begin{figure}[t] 
    \centering
    \includegraphics[width=0.9\linewidth]{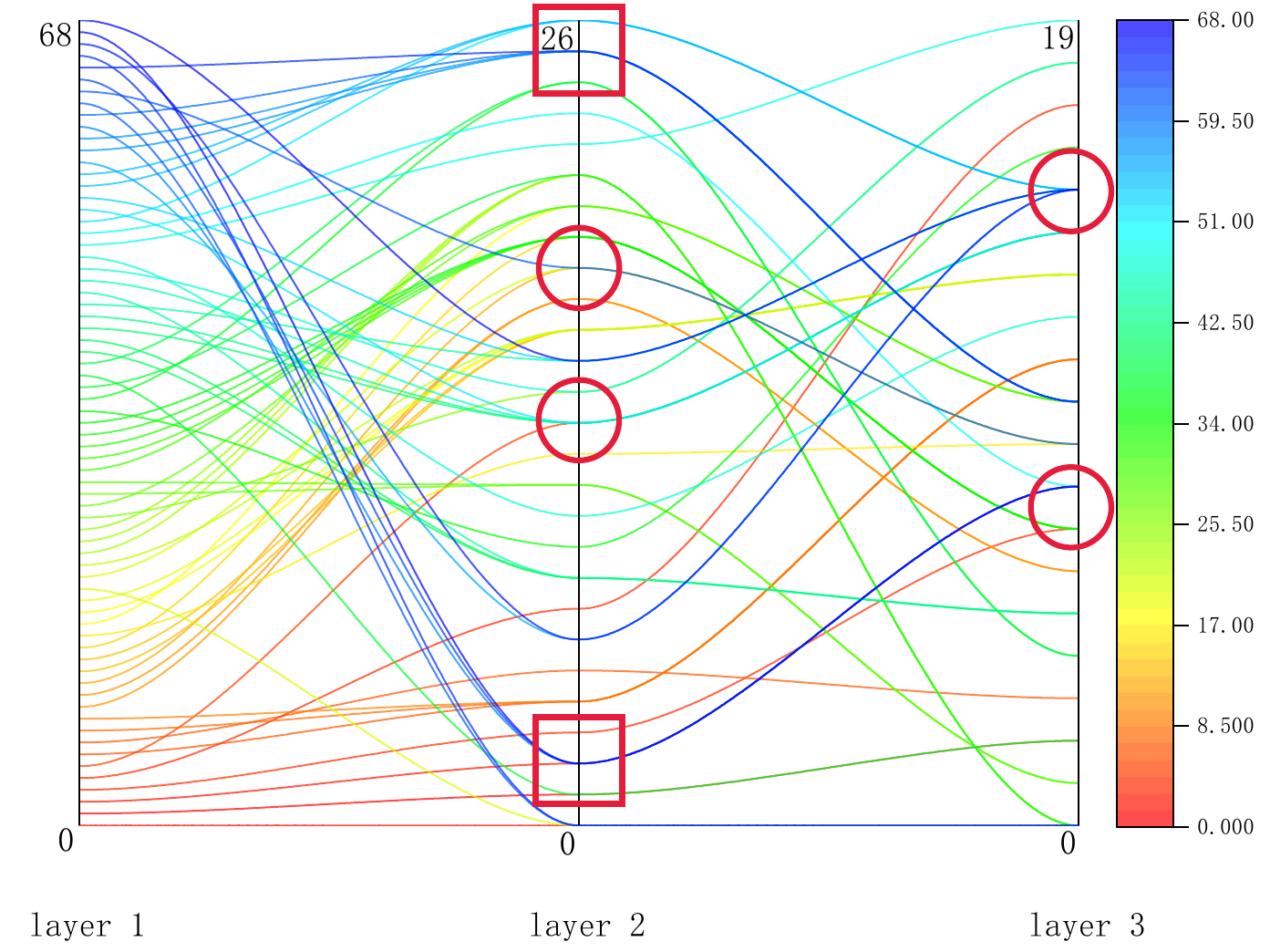}
    \caption{An example of community summarization process in our graph building from QuALITY. Layer 1 consists of 69 small documents, Layer 2 is the community summary (0-26) derived from Layer 1, and Layer 3 is the community summary (0-19) derived from Layer 2.
    }
    \label{fig:layers}
\end{figure}

\textbf{Case Study.} 
Figure~\ref{fig:layers} illustrates a real example from QuALITY, where each line represents the direction in which each point gathers from left to right. From this figure, we can see that we can see that on the second and third layers, whether they were previously far apart or close together, they will gather together according to the community. This indicates that our approach can effectively organize and understand external knowledge, meaning that our summarization-based graph structure is effective. This corresponding textual content, comparison of output results from different models (Sec.~\ref{case_study2}), and comprehensive comparison of graph construction using different methods (Sec.~\ref{case_study3}) can be found in the appendix.
\section{Conclusion}
This work explores a summarization-based hierarchical graph to comprehensively extract both low- and high-level information from external knowledge for answering questions (QA). Moreover, we utilize small-scale LLMs to automatically synthesize data on human preferences and employ mode-seeking loss to capture the main patterns of the optimal policy, thereby better achieving the output of human preferences.
\section{Acknowledgements}
This research/project is supported by the Ministry of Education, Singapore under its MOE Academic Research Fund Tier 2 (MOE-T2EP20123-0005: Neurosymbolic AI for Commonsense-based Question Answering in Multiple Domains, by the National Natural Science Foundation of China grant (NSFC No.~62206193 and No.~62376178), and by the General Research Fund (GRF) project sponsored by the Research Grants Council Hong Kong (Project No.~15611021). 
\section{Limitations}
This study has several limitations, primarily due to computational resource constraints. Specifically, we were unable to utilize LLMs with more than 8 billion parameters or state-of-the-art (SOTA) models such as GPT-4. Consequently, some prior studies could not be fully reproduced. Nevertheless, our method achieves superior performance on several benchmark datasets compared to approaches leveraging GPT-4 as the backbone model, demonstrating its effectiveness despite these limitations.

\bibliography{main}

\begin{thebibliography}{44}
\providecommand{\natexlab}[1]{#1}

\bibitem[{Alwaneen et~al.(2022)Alwaneen, Azmi, Aboalsamh, Cambria, and
  Hussain}]{alwara}
Tahani~H Alwaneen, Aqil~M Azmi, Hatim~A Aboalsamh, Erik Cambria, and Amir
  Hussain. 2022.
\newblock Arabic question answering system: A survey.
\newblock \emph{Artificial Intelligence Review}, 55(1):207--253.

\bibitem[{Asai et~al.(2024)Asai, Wu, Wang, Sil, and
  Hajishirzi}]{asaiSELFRAGLEARNINGRETRIEVE2024}
Akari Asai, Zeqiu Wu, Yizhong Wang, Avirup Sil, and Hannaneh Hajishirzi. 2024.
\newblock {{SELF-RAG}}: {{LEARNING TO RETRIEVE}}, {{GENERATE}}, {{AND CRITIQUE
  THROUGH SELF-REFLECTION}}.

\bibitem[{Barnett et~al.(2024)Barnett, Kurniawan, Thudumu, Brannelly, and
  Abdelrazek}]{barnett2024sevenfailurepointsengineering}
Scott Barnett, Stefanus Kurniawan, Srikanth Thudumu, Zach Brannelly, and
  Mohamed Abdelrazek. 2024.
\newblock \href {https://arxiv.org/abs/2401.05856} {Seven failure points when
  engineering a retrieval augmented generation system}.
\newblock \emph{Preprint}, arXiv:2401.05856.

\bibitem[{Cambria et~al.(2013)Cambria, Schuller, Liu, Wang, and
  Havasi}]{camsta}
Erik Cambria, Bjoern Schuller, Bing Liu, Haixun Wang, and Catherine Havasi.
  2013.
\newblock Statistical approaches to concept-level sentiment analysis.
\newblock \emph{{IEEE} Intelligent Systems}, 28(3):6--9.

\bibitem[{Chan et~al.(2022)Chan, Silva, Lim, Kozuno, Mahmood, and
  White}]{jmlr/0001SLKMW22}
Alan Chan, Hugo Silva, Sungsu Lim, Tadashi Kozuno, A.~Rupam Mahmood, and Martha
  White. 2022.
\newblock \href {https://jmlr.org/papers/v23/21-054.html} {Greedification
  operators for policy optimization: Investigating forward and reverse {KL}
  divergences}.
\newblock \emph{J. Mach. Learn. Res.}, 23:253:1--253:79.

\bibitem[{Chandler and Sweller(1991)}]{chandler1991cognitive}
Paul Chandler and John Sweller. 1991.
\newblock Cognitive load theory and the format of instruction.
\newblock \emph{Cognition and instruction}, 8(4):293--332.

\bibitem[{Chen et~al.(2024)Chen, Xiao, Zhang, Luo, Lian, and
  Liu}]{chen2024bgem3embeddingmultilingualmultifunctionality}
Jianlv Chen, Shitao Xiao, Peitian Zhang, Kun Luo, Defu Lian, and Zheng Liu.
  2024.
\newblock \href {https://arxiv.org/abs/2402.03216} {Bge m3-embedding:
  Multi-lingual, multi-functionality, multi-granularity text embeddings through
  self-knowledge distillation}.
\newblock \emph{Preprint}, arXiv:2402.03216.

\bibitem[{Christiano et~al.(2023)Christiano, Leike, Brown, Martic, Legg, and
  Amodei}]{christiano2023deepreinforcementlearninghuman}
Paul Christiano, Jan Leike, Tom~B. Brown, Miljan Martic, Shane Legg, and Dario
  Amodei. 2023.
\newblock \href {https://arxiv.org/abs/1706.03741} {Deep reinforcement learning
  from human preferences}.
\newblock \emph{Preprint}, arXiv:1706.03741.

\bibitem[{Dasigi et~al.(2021)Dasigi, Lo, Beltagy, Cohan, Smith, and
  Gardner}]{dasigi2021datasetinformationseekingquestionsanswers}
Pradeep Dasigi, Kyle Lo, Iz~Beltagy, Arman Cohan, Noah~A. Smith, and Matt
  Gardner. 2021.
\newblock \href {https://arxiv.org/abs/2105.03011} {A dataset of
  information-seeking questions and answers anchored in research papers}.
\newblock \emph{Preprint}, arXiv:2105.03011.

\bibitem[{Edge et~al.(2024)Edge, Trinh, Cheng, Bradley, Chao, Mody, Truitt, and
  Larson}]{edgeLocalGlobalGraph2024}
Darren Edge, Ha~Trinh, Newman Cheng, Joshua Bradley, Alex Chao, Apurva Mody,
  Steven Truitt, and Jonathan Larson. 2024.
\newblock \href {https://arxiv.org/abs/2404.16130} {From {{Local}} to
  {{Global}}: {{A Graph RAG Approach}} to {{Query-Focused Summarization}}}.
\newblock \emph{Preprint}, arXiv:2404.16130.

\bibitem[{Ganesan(2018)}]{ganesan2018rouge20updatedimproved}
Kavita Ganesan. 2018.
\newblock \href {https://arxiv.org/abs/1803.01937} {Rouge 2.0: Updated and
  improved measures for evaluation of summarization tasks}.
\newblock \emph{Preprint}, arXiv:1803.01937.

\bibitem[{Gao et~al.(2024)Gao, Xiong, Gao, Jia, Pan, Bi, Dai, Sun, Wang, and
  Wang}]{gaoRetrievalAugmentedGenerationLarge2024}
Yunfan Gao, Yun Xiong, Xinyu Gao, Kangxiang Jia, Jinliu Pan, Yuxi Bi, Yi~Dai,
  Jiawei Sun, Meng Wang, and Haofen Wang. 2024.
\newblock \href {https://arxiv.org/abs/2312.10997} {Retrieval-{{Augmented
  Generation}} for {{Large Language Models}}: {{A Survey}}}.
\newblock \emph{Preprint}, arXiv:2312.10997.

\bibitem[{Grattafiori et~al.(2024)Grattafiori, Dubey, and
  Jauhri}]{grattafiori2024llama3herdmodels}
Aaron Grattafiori, Abhimanyu Dubey, and Abhinav Jauhri. 2024.
\newblock \href {https://arxiv.org/abs/2407.21783} {The llama 3 herd of
  models}.
\newblock \emph{Preprint}, arXiv:2407.21783.

\bibitem[{Guo et~al.(2024)Guo, Xia, Yu, Ao, and Huang}]{guo_lightrag_2024}
Zirui Guo, Lianghao Xia, Yanhua Yu, Tu~Ao, and Chao Huang. 2024.
\newblock \href {https://doi.org/10.48550/arXiv.2410.05779} {{LightRAG}:
  {Simple} and {Fast} {Retrieval}-{Augmented} {Generation}}.
\newblock \emph{arXiv preprint}.
\newblock ArXiv:2410.05779 [cs].

\bibitem[{Huang et~al.(2025)Huang, Mao, Wu, He, Zhang, and Cambria}]{huafun}
Zihao Huang, Rui Mao, Xiaobao Wu, Kai He, Xulang Zhang, and Erik Cambria. 2025.
\newblock Fundamental algorithms for retrieval augmented generation: A survey.
\newblock \emph{IEEE Transactions on Knowledge and Data Engineering}.

\bibitem[{Ji et~al.(2024)Ji, Lu, Niu, Ke, Wang, Zhu, Tang, and
  Huang}]{ji_towards_2024}
Haozhe Ji, Cheng Lu, Yilin Niu, Pei Ke, Hongning Wang, Jun Zhu, Jie Tang, and
  Minlie Huang. 2024.
\newblock Towards efficient exact optimization of language model alignment.
\newblock In \emph{Proceedings of the 41st International Conference on Machine
  Learning}, ICML'24. JMLR.org.

\bibitem[{Jin et~al.(2020)Jin, Pan, Oufattole, Weng, Fang, and
  Szolovits}]{jin2020diseasedoespatienthaveMedQA}
Di~Jin, Eileen Pan, Nassim Oufattole, Wei-Hung Weng, Hanyi Fang, and Peter
  Szolovits. 2020.
\newblock \href {https://arxiv.org/abs/2009.13081} {What disease does this
  patient have? a large-scale open domain question answering dataset from
  medical exams}.
\newblock \emph{Preprint}, arXiv:2009.13081.

\bibitem[{Jin et~al.(2019)Jin, Dhingra, Liu, Cohen, and
  Lu}]{jin2019pubmedqadatasetbiomedicalresearch}
Qiao Jin, Bhuwan Dhingra, Zhengping Liu, William~W. Cohen, and Xinghua Lu.
  2019.
\newblock \href {https://arxiv.org/abs/1909.06146} {Pubmedqa: A dataset for
  biomedical research question answering}.
\newblock \emph{Preprint}, arXiv:1909.06146.

\bibitem[{Lewis et~al.(2021)Lewis, Perez, Piktus, Petroni, Karpukhin, Goyal,
  K{\"u}ttler, Lewis, Yih, Rockt{\"a}schel, Riedel, and
  Kiela}]{lewisRetrievalAugmentedGenerationKnowledgeIntensive2021}
Patrick Lewis, Ethan Perez, Aleksandra Piktus, Fabio Petroni, Vladimir
  Karpukhin, Naman Goyal, Heinrich K{\"u}ttler, Mike Lewis, Wen-tau Yih, Tim
  Rockt{\"a}schel, Sebastian Riedel, and Douwe Kiela. 2021.
\newblock \href {https://arxiv.org/abs/2005.11401} {Retrieval-{{Augmented
  Generation}} for {{Knowledge-Intensive NLP Tasks}}}.
\newblock \emph{Preprint}, arXiv:2005.11401.

\bibitem[{Lin et~al.(2021{\natexlab{a}})Lin, Wu, Yang, Lee, and
  Ren}]{lin2021riddlesensereasoningriddlequestions}
Bill~Yuchen Lin, Ziyi Wu, Yichi Yang, Dong-Ho Lee, and Xiang Ren.
  2021{\natexlab{a}}.
\newblock \href {https://arxiv.org/abs/2101.00376} {Riddlesense: Reasoning
  about riddle questions featuring linguistic creativity and commonsense
  knowledge}.
\newblock \emph{Preprint}, arXiv:2101.00376.

\bibitem[{Lin(2004)}]{lin-2004-rouge}
Chin-Yew Lin. 2004.
\newblock \href {https://aclanthology.org/W04-1013} {{ROUGE}: A package for
  automatic evaluation of summaries}.
\newblock In \emph{Text Summarization Branches Out}, pages 74--81, Barcelona,
  Spain. Association for Computational Linguistics.

\bibitem[{Lin et~al.(2021{\natexlab{b}})Lin, Jaech, Li, Gormley, and
  Eisner}]{naacl/LinJLGE21}
Chu{-}Cheng Lin, Aaron Jaech, Xin Li, Matthew~R. Gormley, and Jason Eisner.
  2021{\natexlab{b}}.
\newblock \href {https://doi.org/10.18653/V1/2021.NAACL-MAIN.405} {Limitations
  of autoregressive models and their alternatives}.
\newblock In \emph{Proceedings of the 2021 Conference of the North American
  Chapter of the Association for Computational Linguistics: Human Language
  Technologies, {NAACL-HLT} 2021, Online, June 6-11, 2021}, pages 5147--5173.
  Association for Computational Linguistics.

\bibitem[{Liu et~al.(2023)Liu, Lin, Hewitt, Paranjape, Bevilacqua, Petroni, and
  Liang}]{liu2023lostmiddlelanguagemodels}
Nelson~F. Liu, Kevin Lin, John Hewitt, Ashwin Paranjape, Michele Bevilacqua,
  Fabio Petroni, and Percy Liang. 2023.
\newblock \href {https://arxiv.org/abs/2307.03172} {Lost in the middle: How
  language models use long contexts}.
\newblock \emph{Preprint}, arXiv:2307.03172.

\bibitem[{Liu et~al.(2024)Liu, Geng, Wang, Cambria, and Jiang}]{liudis}
Qian Liu, Xiubo Geng, Yu~Wang, Erik Cambria, and Daxin Jiang. 2024.
\newblock Disentangled retrieval and reasoning for implicit question answering.
\newblock \emph{{IEEE} Transactions on Neural Networks and Learning Systems},
  35(6):7804--7815.

\bibitem[{{Mistral AI}(2025)}]{ministraux_update}
{Mistral AI}. 2025.
\newblock {Un Ministral, des Ministraux}.
\newblock \url{https://mistral.ai/news/ministraux/}.
\newblock Accessed: 2025-01-19.

\bibitem[{Nguyen et~al.(2024)Nguyen, Chin, and
  Tai}]{nguyenRewardRAGEnhancingRAG2024}
Thang Nguyen, Peter Chin, and Yu-Wing Tai. 2024.
\newblock \href {https://arxiv.org/abs/2410.03780} {Reward-{{RAG}}: {{Enhancing
  RAG}} with {{Reward Driven Supervision}}}.
\newblock \emph{Preprint}, arXiv:2410.03780.

\bibitem[{Oneto et~al.(2016)Oneto, Bisio, Cambria, and Anguita}]{onesta}
Luca Oneto, Federica Bisio, Erik Cambria, and Davide Anguita. 2016.
\newblock Statistical learning theory and {ELM} for big social data analysis.
\newblock \emph{{IEEE} Computational Intelligence Magazine}, 11(3):45--55.

\bibitem[{OpenAI et~al.(2024)OpenAI, Achiam, Adler, Agarwal, and
  Ahmad}]{openai2024gpt4technicalreport}
OpenAI, Josh Achiam, Steven Adler, Sandhini Agarwal, and Lama Ahmad. 2024.
\newblock \href {https://arxiv.org/abs/2303.08774} {Gpt-4 technical report}.
\newblock \emph{Preprint}, arXiv:2303.08774.

\bibitem[{Ouyang et~al.(2022)Ouyang, Wu, Jiang, Almeida, Wainwright, Mishkin,
  Zhang, Agarwal, Slama, Ray, Schulman, Hilton, Kelton, Miller, Simens, Askell,
  Welinder, Christiano, Leike, and
  Lowe}]{ouyang2022traininglanguagemodelsfollow}
Long Ouyang, Jeff Wu, Xu~Jiang, Diogo Almeida, Carroll~L. Wainwright, Pamela
  Mishkin, Chong Zhang, Sandhini Agarwal, Katarina Slama, Alex Ray, John
  Schulman, Jacob Hilton, Fraser Kelton, Luke Miller, Maddie Simens, Amanda
  Askell, Peter Welinder, Paul Christiano, Jan Leike, and Ryan Lowe. 2022.
\newblock \href {https://arxiv.org/abs/2203.02155} {Training language models to
  follow instructions with human feedback}.
\newblock \emph{Preprint}, arXiv:2203.02155.

\bibitem[{Pal et~al.(2022)Pal, Umapathi, and
  Sankarasubbu}]{pal2022medmcqalargescalemultisubject}
Ankit Pal, Logesh~Kumar Umapathi, and Malaikannan Sankarasubbu. 2022.
\newblock \href {https://arxiv.org/abs/2203.14371} {Medmcqa : A large-scale
  multi-subject multi-choice dataset for medical domain question answering}.
\newblock \emph{Preprint}, arXiv:2203.14371.

\bibitem[{Pang et~al.(2022)Pang, Parrish, Joshi, Nangia, Phang, Chen,
  Padmakumar, Ma, Thompson, He, and
  Bowman}]{pang2022qualityquestionansweringlong}
Richard~Yuanzhe Pang, Alicia Parrish, Nitish Joshi, Nikita Nangia, Jason Phang,
  Angelica Chen, Vishakh Padmakumar, Johnny Ma, Jana Thompson, He~He, and
  Samuel~R. Bowman. 2022.
\newblock \href {https://arxiv.org/abs/2112.08608} {Quality: Question answering
  with long input texts, yes!}
\newblock \emph{Preprint}, arXiv:2112.08608.

\bibitem[{Patterson et~al.(2022)Patterson, Gonzalez, H{\"o}lzle, Le, Liang,
  Munguia, Rothchild, So, Texier, and
  Dean}]{patterson2022carbonfootprintmachinelearning}
David Patterson, Joseph Gonzalez, Urs H{\"o}lzle, Quoc Le, Chen Liang,
  Lluis-Miquel Munguia, Daniel Rothchild, David So, Maud Texier, and Jeff Dean.
  2022.
\newblock \href {https://arxiv.org/abs/2204.05149} {The carbon footprint of
  machine learning training will plateau, then shrink}.
\newblock \emph{Preprint}, arXiv:2204.05149.

\bibitem[{Qwen et~al.(2025)Qwen, :, Yang, Yang, Zhang, Hui, Zheng, Yu, Li, Liu,
  Huang, Wei, Lin, Yang, Tu, Zhang, Yang, Yang, Zhou, Lin, Dang, Lu, Bao, Yang,
  Yu, Li, Xue, Zhang, Zhu, Men, Lin, Li, Tang, Xia, Ren, Ren, Fan, Su, Zhang,
  Wan, Liu, Cui, Zhang, and Qiu}]{qwen2025qwen25technicalreport}
Qwen, :, An~Yang, Baosong Yang, Beichen Zhang, Binyuan Hui, Bo~Zheng, Bowen Yu,
  Chengyuan Li, Dayiheng Liu, Fei Huang, Haoran Wei, Huan Lin, Jian Yang,
  Jianhong Tu, Jianwei Zhang, Jianxin Yang, Jiaxi Yang, Jingren Zhou, Junyang
  Lin, Kai Dang, Keming Lu, Keqin Bao, Kexin Yang, Le~Yu, Mei Li, Mingfeng Xue,
  Pei Zhang, Qin Zhu, Rui Men, Runji Lin, Tianhao Li, Tianyi Tang, Tingyu Xia,
  Xingzhang Ren, Xuancheng Ren, Yang Fan, Yang Su, Yichang Zhang, Yu~Wan,
  Yuqiong Liu, Zeyu Cui, Zhenru Zhang, and Zihan Qiu. 2025.
\newblock \href {https://arxiv.org/abs/2412.15115} {Qwen2.5 technical report}.
\newblock \emph{Preprint}, arXiv:2412.15115.

\bibitem[{Rafailov et~al.(2024)Rafailov, Sharma, Mitchell, Ermon, Manning, and
  Finn}]{rafailov2024directpreferenceoptimizationlanguage}
Rafael Rafailov, Archit Sharma, Eric Mitchell, Stefano Ermon, Christopher~D.
  Manning, and Chelsea Finn. 2024.
\newblock \href {https://arxiv.org/abs/2305.18290} {Direct preference
  optimization: Your language model is secretly a reward model}.
\newblock \emph{Preprint}, arXiv:2305.18290.

\bibitem[{Sarthi et~al.(2024)Sarthi, Abdullah, Tuli, Khanna, Goldie, and
  Manning}]{sarthiRAPTORRecursiveAbstractive2024}
Parth Sarthi, Salman Abdullah, Aditi Tuli, Shubh Khanna, Anna Goldie, and
  Christopher~D. Manning. 2024.
\newblock \href {https://arxiv.org/abs/2401.18059} {{{RAPTOR}}: {{Recursive
  Abstractive Processing}} for {{Tree-Organized Retrieval}}}.
\newblock \emph{Preprint}, arXiv:2401.18059.

\bibitem[{Sweller(1988)}]{SWELLER1988257}
John Sweller. 1988.
\newblock \href {https://doi.org/10.1016/0364-0213(88)90023-7} {Cognitive load
  during problem solving: Effects on learning}.
\newblock \emph{Cognitive Science}, 12(2):257--285.

\bibitem[{Tang et~al.(2024)Tang, Zhao, Zhu, Xiao, Kasikci, and
  Han}]{tang2024questqueryawaresparsityefficient}
Jiaming Tang, Yilong Zhao, Kan Zhu, Guangxuan Xiao, Baris Kasikci, and Song
  Han. 2024.
\newblock \href {https://arxiv.org/abs/2406.10774} {Quest: Query-aware sparsity
  for efficient long-context llm inference}.
\newblock \emph{Preprint}, arXiv:2406.10774.

\bibitem[{Traag et~al.(2019)Traag, Waltman, and Van~Eck}]{traag2019louvain}
Vincent~A Traag, Ludo Waltman, and Nees~Jan Van~Eck. 2019.
\newblock From louvain to leiden: guaranteeing well-connected communities.
\newblock \emph{Scientific reports}, 9(1):1--12.

\bibitem[{Wang et~al.(2024)Wang, Yang, Huang, Yang, Majumder, and
  Wei}]{wang-etal-2024-improving-text_gpt_gen_data}
Liang Wang, Nan Yang, Xiaolong Huang, Linjun Yang, Rangan Majumder, and Furu
  Wei. 2024.
\newblock \href {https://doi.org/10.18653/v1/2024.acl-long.642} {Improving text
  embeddings with large language models}.
\newblock In \emph{Proceedings of the 62nd Annual Meeting of the Association
  for Computational Linguistics (Volume 1: Long Papers)}, pages 11897--11916,
  Bangkok, Thailand. Association for Computational Linguistics.

\bibitem[{Wei et~al.(2023)Wei, Wang, Schuurmans, Bosma, Ichter, Xia, Chi, Le,
  and Zhou}]{wei2023chainofthoughtpromptingelicitsreasoning}
Jason Wei, Xuezhi Wang, Dale Schuurmans, Maarten Bosma, Brian Ichter, Fei Xia,
  Ed~Chi, Quoc Le, and Denny Zhou. 2023.
\newblock \href {https://arxiv.org/abs/2201.11903} {Chain-of-thought prompting
  elicits reasoning in large language models}.
\newblock \emph{Preprint}, arXiv:2201.11903.

\bibitem[{Wu et~al.(2024)Wu, Chen, Deng, Sabour, Meng, and
  Huang}]{acl/Wu0DSMH24}
Jincenzi Wu, Zhuang Chen, Jiawen Deng, Sahand Sabour, Helen Meng, and Minlie
  Huang. 2024.
\newblock \href {https://doi.org/10.18653/V1/2024.ACL-LONG.848} {{COKE:} {A}
  cognitive knowledge graph for machine theory of mind}.
\newblock In \emph{Proceedings of the 62nd Annual Meeting of the Association
  for Computational Linguistics (Volume 1: Long Papers), {ACL} 2024, Bangkok,
  Thailand, August 11-16, 2024}, pages 15984--16007. Association for
  Computational Linguistics.

\bibitem[{Xiong et~al.(2024)Xiong, Jin, Lu, and
  Zhang}]{xiong2024benchmarkingretrievalaugmentedgenerationmedicineMirage}
Guangzhi Xiong, Qiao Jin, Zhiyong Lu, and Aidong Zhang. 2024.
\newblock \href {https://arxiv.org/abs/2402.13178} {Benchmarking
  retrieval-augmented generation for medicine}.
\newblock \emph{Preprint}, arXiv:2402.13178.

\bibitem[{Xu et~al.(2024)Xu, Pang, Yu, Meng, Shen, Cheng, and
  Zhou}]{xuUnsupervisedInformationRefinement2024}
Shicheng Xu, Liang Pang, Mo~Yu, Fandong Meng, Huawei Shen, Xueqi Cheng, and Jie
  Zhou. 2024.
\newblock \href {https://doi.org/10.18653/v1/2024.acl-long.9} {Unsupervised
  {{Information Refinement Training}} of {{Large Language Models}} for
  {{Retrieval-Augmented Generation}}}.
\newblock In \emph{Proceedings of the 62nd {{Annual Meeting}} of the
  {{Association}} for {{Computational Linguistics}} ({{Volume}} 1: {{Long
  Papers}})}, pages 133--145. Association for Computational Linguistics.

\bibitem[{Zhao et~al.(2024)Zhao, Zhang, Yu, Wang, Geng, Fu, Yang, Zhang, Jiang,
  and Cui}]{zhao2024retrievalaugmentedgenerationaigeneratedcontent}
Penghao Zhao, Hailin Zhang, Qinhan Yu, Zhengren Wang, Yunteng Geng, Fangcheng
  Fu, Ling Yang, Wentao Zhang, Jie Jiang, and Bin Cui. 2024.
\newblock \href {https://arxiv.org/abs/2402.19473} {Retrieval-augmented
  generation for ai-generated content: A survey}.
\newblock \emph{Preprint}, arXiv:2402.19473.

\end{thebibliography}
\newpage
\appendix
\begin{figure*}[t!]
    \centering
    \includegraphics[width=1\linewidth]{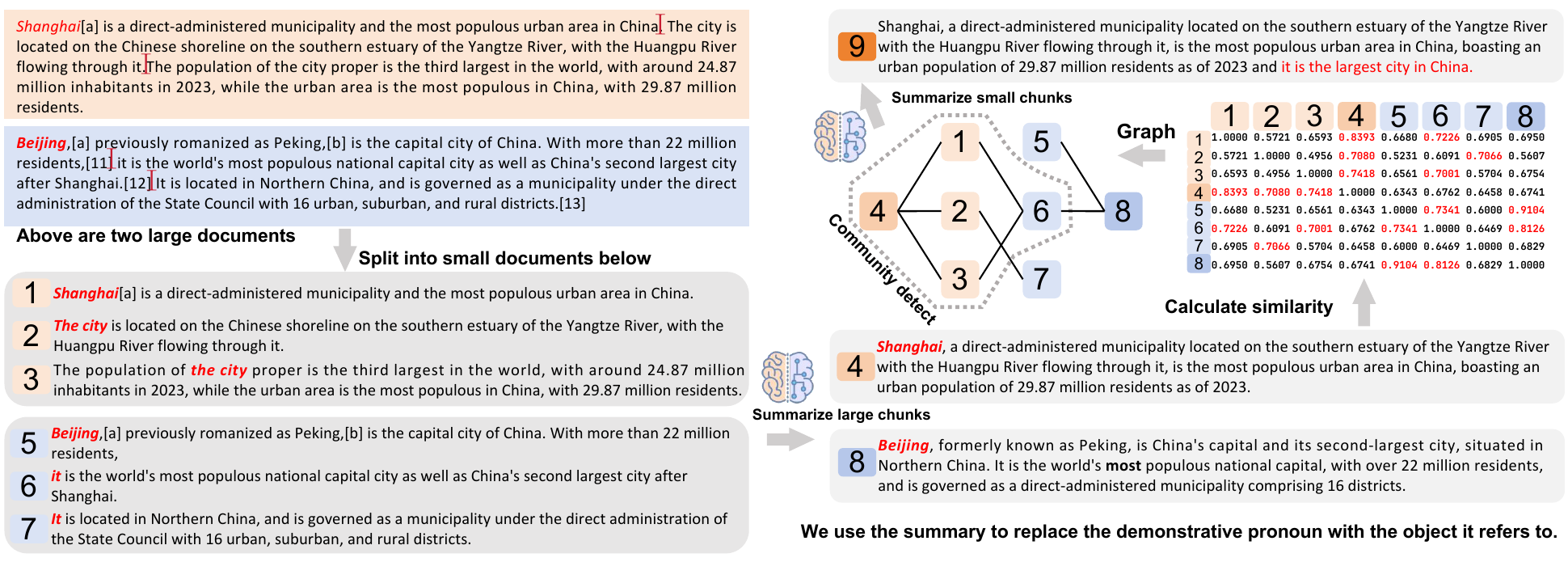}
    \caption{The figure shows two descriptions of the cities 
    \sethlcolor{shanghai}\hl{Shanghai} and 
    \sethlcolor{beijing}\hl{Beijing} form wikipedia.
    As part of the basic RAG process, we split the large documents into 
    \sethlcolor{chunk}\hl{small chunks (documents)} 
    \sethlcolor{shanghai}\hl{1 \textasciitilde 3} 
    \sethlcolor{beijing}\hl{5 \textasciitilde 7}, embed them and store them in a vector database. During retrieval, the basic RAG process may return incorrect references (Query "\sethlcolor{shanghai}\hl{Shanghai}" get \sethlcolor{beijing}\hl{6}). To address this, we apply abstractive summarization to large documents to eliminate misrepresentation. Furthermore, we build a graph $\mathcal{G}$
    based on the similarity between documents. Using this graph, we group similar documents into the same community and apply abstractive summarization to obtain broader insights.
    }
    \label{fig:intro_2}
\end{figure*}
\section{A Simple Example of Graph Construction}

Figure~\ref{fig:intro_2} illustrates an example of graph construction provided using two documents about cities in China, sourced from Wikipedia. First, we summarize each document. Next, we split the documents into smaller segments. Then, we embed these segments into dense vectors and calculate the similarity between them. We treat the segments as nodes, and based on their similarity, we establish edges between the nodes. Using this graph, we apply community detection algorithms to identify communities, which are then summarized to extract high-level nodes.
\section{Details of Mode-seeking Preference Alignment}
We provide a simple justification for why our mode-seeking loss with small-scale auto-synthesized training data is more satisfying than the mean-seeking loss of traditional DPO with large-scale LLM-generated training data.
Optimal Policy $\pi^*$ is defined as follows:

\begin{equation}
\begin{split}
\pi^{*}&=\arg \max _{\pi} \mathbb{E}_{x \sim D, y \sim \pi(y | x)}\left[r_{\phi}(x, y)\right] \\
&-\beta \cdot D_{K L}\left(\pi(y | x) \| \pi_{\mathrm{sft}}(y | x)\right)
\end{split}
\end{equation}
where $r_{\phi}(x, y)$ is the reward model, $\pi_{\mathrm{sft}}(y | x)$ is the initial policy, and $\beta$ is a regularization parameter.
MS minimizes the reverse KL divergence:
\begin{equation}
\pi_{ms}=\arg \max _\pi D_{KL}(\pi(y |x) || \pi^*(y |x))
\end{equation}
DPO minimizes the forward KL divergence:
\begin{equation}
\pi_{dpo}=\arg \max _\pi D_{KL}(\pi^*(y |x) || \pi(y |x))
\end{equation}
The reverse KL divergence is defined as follows:
\begin{equation}
\begin{split}
D_{K L}\left(\pi(y \mid x) \| \pi^{*}(y \mid x)\right) \\=\mathbb{E}_{y \sim \pi(y \mid x)}\left[\log \frac{\pi(y \mid x)}{\pi^{*}(y \mid x)}\right]
\end{split}
\end{equation}
Mode-Seeking: Minimizing reverse KL divergence $D_{KL}\left(\pi(y \mid x) \| \pi^{*}(y \mid x)\right)$  encourages $\pi(y \mid x)$ to concentrate on the main modes of $\pi^*(y \mid x)$.
The forward KL divergence is defined as follows:
\begin{equation}
\begin{split}
D_{K L}\left(\pi^{*}(y \mid x) \| \pi(y \mid x)\right) \\=\mathbb{E}_{y \sim \pi(y \mid x)}\left[\log \frac{\pi^{*}(y \mid x)}{\pi(y \mid x)}\right]
\end{split}
\end{equation}

Mean-Seeking: Minimizing forward KL divergence $D_{K L}\left(\pi^{*}(y \mid x) \| \pi(y \mid x)\right)$ encourages $\pi^*(y \mid x)$ to cover the entire support of $\pi^*(y \mid x)$, potentially leading to a distribution that is less concentrated on the main modes~\cite{ji_towards_2024}.
Minimizing the reverse KL divergence $D_{KL}(\pi(y |x) || \pi^*(y |x))$ causes $\pi_{ms}$ to concentrate on the main modes of $\pi^*$ because it penalizes $\pi_{ms}$ for assigning high probability to regions where $\pi^*$ has low probability. This mode-seeking behavior ensures that $\pi_{ms}$ closely approximates the main modes of $\pi^*$, making it an effective way to optimize policies in alignment tasks.
Simply put, building on the success of Direct Preference Optimization (DPO)~\cite{rafailov2024directpreferenceoptimizationlanguage} based on reinforcement learning in the training of LLMs, we propose to seek the mode distribution of the optimal policy~\cite{ji_towards_2024} with the mode-seeking loss function (MS) as follows:

\begin{equation}
\begin{split}
\mathcal{L}_{\text{ms}}(\pi_\theta) &= \mathbb{E}_{x \sim \mathcal{D}_{\text{pref}}} \mathbb{E}_{\pi_{\text{sft}}}(\mathbf{y}_{1:K} | x) \Big[ \\
& \mathbb{D}_{\text{KL}}(p_{f_\theta}(\cdot | \mathbf{y}_{1:K}, x) \| p_{r_\phi}(\cdot | \mathbf{y}_{1:K}, x)) \Big]
\end{split}
\end{equation}
where $\mathcal{D}_{\text{pref}}$ is the preference dataset which contains human-labeled preference data.  Each entry includes a prompt $x$ and a set of responses $(y_{1:k})$, where $y_1,...y_k$ are sampled responses and Labels indicating which responses are preferred or rejected.
$\pi_{\text{sft}}(y_{1:K}|x)$ is the supervised fine-tuned (SFT) policy $\pi_{\text{sft}}$ is the probability distribution of responses given a prompt $x$ after the language model has undergone supervised fine-tuning.
$p_{f_\theta}(\cdot | y_{1:K}, x)$ is the empirical distribution based on the model policy. This distribution is computed over the sampled responses $y_{1:K}$ and reflects the model's current policy $\pi_\theta$. Represents the model's belief over the sampled responses $y_{1:K}$.
$p_{r_\phi}(\cdot | y_{1:K}, x)$is the empirical distribution based on the reward model. This distribution reflects the reward model's scoring over the sampled responses $y_{1:K}$, based on human preferences. Represents the "ideal" distribution of responses based on human preferences.
$\mathbb{D}_{\text{KL}}(p_{f_\theta} \| p_{r_\phi})$: Kullback-Leibler (KL) divergence, which aligns the model's output distribution $p_{f_\theta}$ with the distribution defined by the reward model $p_{r_\phi}$.
$\mathbb{E}_{x \sim \mathcal{D}_{\text{pref}}}$ ensures generalization of the policy $\pi_\theta$ to the entire dataset.
$\mathbb{E}_{\pi_{\text{sft}}(y_{1:K}|x)}$ captures the effect of sampling different response sets on the loss.
The MS loss aligns the model's policy $\pi_\theta$ with human preferences by minimizing the KL divergence between the model's empirical distribution ($p_{f_\theta}$) and the reward model's empirical distribution ($p_{r_\phi}$) over sampled responses, averaged across all prompts in the dataset. It effectively optimizes the policy while maintaining computational efficiency.
\section{Details of Baselines}
\subsection{Baselines}
\begin{itemize}
\item Standard RAG~\cite{lewisRetrievalAugmentedGenerationKnowledgeIntensive2021,
gaoRetrievalAugmentedGenerationLarge2024} is the most classic method for retrieving external knowledge to augment the performance of LLMs.
\item RAPTOR~\cite{sarthiRAPTORRecursiveAbstractive2024} builds a summary tree with text chunk embedding and clustering and retrieves from the tree to offer LLM additional information from the outer.
\item LightRAG~\cite{guo_lightrag_2024} incorporates graph structures into text indexing and retrieval processes and allows the system to remain effective and responsive in rapidly changing data environments compared to the Graph RAG~\cite{edgeLocalGlobalGraph2024} which builds a graph using LLM to extract entity-relation triples form text and makes summaries form that to build a higher-level graph, then retrieves from these graphs to provide LLM outer knowledge.
\item Reward-RAG~\cite{nguyenRewardRAGEnhancingRAG2024} aligns RAG with human preferences by integrating a reward model to train a higher-performing embedding model.
\end{itemize}
\section{Details of Datasets}
\subsection{Datasets}
QASPER~\cite{dasigi2021datasetinformationseekingquestionsanswers} 
consists of 5,049 questions over 1,585 Natural Language Processing papers that are often read to seek information present in the full text to answer specific questions.
QuALITY~\cite{pang2022qualityquestionansweringlong},
Question Answering with Long Input Texts, Yes! is a dataset to enable training and testing models on long-document comprehension, which consists of multiple-choice QA with context passages in English.
RiddleSense~\cite{lin2021riddlesensereasoningriddlequestions}
consists of 5.7k examples and aims to build and test the model's reasoning about riddle questions consisting of questions and multiple choice.
PubMedQA~\cite{jin2019pubmedqadatasetbiomedicalresearch}
is a biomedical question-answering (QA) dataset collected from PubMed abstracts that have 1k expert annotations. The task of PubMedQA is to answer research questions with yes/no/maybe.
MedQA~\cite{jin2020diseasedoespatienthaveMedQA} 
collected from the professional medical board exams, which contain 12,723 questions in English, and each question is accompanied by its answer and several options with an alpha index.   
MedMCQA~\cite{pal2022medmcqalargescalemultisubject}
is a large-scale, Multiple-Choice Question Answering (MCQA) dataset built from 194k  high-quality medical exams, designed to address real-world medical entrance exam questions. Each sample contains a question, correct answer(s), and other options.
\section{Details of Evaluation Metrics}
\subsection{Evaluation Metrics} 
We evaluate text generation outputs using the ROUGE score (Recall-Oriented Understudy for Gisting Evaluation)\cite{lin-2004-rouge,ganesan2018rouge20updatedimproved} F1, a standard metric for assessing the quality of generated text by comparing n-gram overlaps between the predicted and reference outputs. For tasks involving discrete responses, such as yes/no, A/B/C/D, or 1/2/3/4 choices, we report accuracy as the evaluation metric.
Additionally, for evaluating our model's performance on medical question-answering datasets, we employ MIRAGE (Medical Information Retrieval-Augmented Generation Evaluation)~\cite{xiong2024benchmarkingretrievalaugmentedgenerationmedicineMirage}
\footnote{\url{https://github.com/mirage-project/mirage}}, 
a specialized metric designed for assessing retrieval-augmented generation models in the medical domain. We apply MIRAGE to evaluate our performance on three benchmark datasets: PubMedQA, MedQA, and MeMCdQA.
\section{Detailed Metrics of QASPER}
\begin{table}[h!]
\centering
\resizebox{\linewidth}{!}{
\begin{tabular}{llllll}	\toprule
         QASPER    & F1     & Precision & Recall & BLEU   & Meteor \\ \hline
\multicolumn{6}{c}{RAPTOR~\cite{sarthiRAPTORRecursiveAbstractive2024}}  \\ \hline                  
\textbf{LLaMa 8B}   & 0.3657 & 0.4660    & 0.3009 & 0.1158 & 0.3338 \\ \hline
\multicolumn{6}{c}{LightGraphRAG~\cite{guo_lightrag_2024}}      \\ \hline
\textbf{LLaMa 8B} & 0.3585 & 0.5781    & 0.2598 & 0.1262 & 0.3834 \\ \hline
\multicolumn{6}{c}{LLaMa 8B~\cite{grattafiori2024llama3herdmodels}}  \\ \hline
\textbf{Basic LLM}           & 0.1040 & 0.1612    & 0.0767 & 0.0221 & 0.0996 \\
\textbf{Basic RAG}           & 0.3599 & 0.5307    & 0.2800 & 0.1723 & 0.3519 \\
\textbf{GraphMPA (ours) }    & 0.3775 & 0.5645    & 0.2835 & 0.1745 & 0.3982 \\ \hline
\multicolumn{6}{c}{Qwen 7B~\cite{qwen2025qwen25technicalreport}}      \\ \hline
\textbf{Basic LLM}           & 0.0881 & 0.1436    & 0.0813 & 0.0174 & 0.0913 \\
\textbf{Basic RAG}           & 0.2654 & 0.6645    & 0.2369 & 0.1675 & 0.3321 \\
\textbf{GraphMPA (ours) }    & 0.3734 & 0.6420    & 0.2632 & 0.1885 & 0.3921 \\ \hline
\multicolumn{6}{c}{Mistral 8B~\cite{ministraux_update}}      \\ \hline
\textbf{Basic LLM}           & 0.1135 & 0.1784    & 0.0832 & 0.0238 & 0.1074 \\
\textbf{Basic RAG}           & 0.3228 & 0.5919    & 0.2219 & 0.1294 & 0.3824 \\
\textbf{GraphMPA (ours) }    & 0.3873 & 0.6314    & 0.2793 & 0.1647 & 0.4135 \\ \bottomrule
\end{tabular}
}
\caption{
The performance of various models on QASPER is evaluated across several metrics, including F1, Precision, Recall, BLEU, and Meteor. We present results for the RAPTOR~\cite{sarthiRAPTORRecursiveAbstractive2024}, LightGraphRAG~\cite{guo_lightrag_2024}, LLaMa 8B~\cite{grattafiori2024llama3herdmodels}, Qwen 7B~\cite{qwen2025qwen25technicalreport}, and Mistral 8B~\cite{ministraux_update} models, with a focus on different configurations: Basic LLM, Basic RAG, and GraphMPA (our approach).
}
\end{table}
Overall, our GraphMPA consistently outperforms all baselines in terms of F1, Precision, Recall, BLEU, and Meteor scores, demonstrating the effectiveness of our approach.
\section{Impact of $\tau$ Value in Graph.}
\begin{figure}[h!]
    \centering
    \includegraphics[width=1\linewidth]{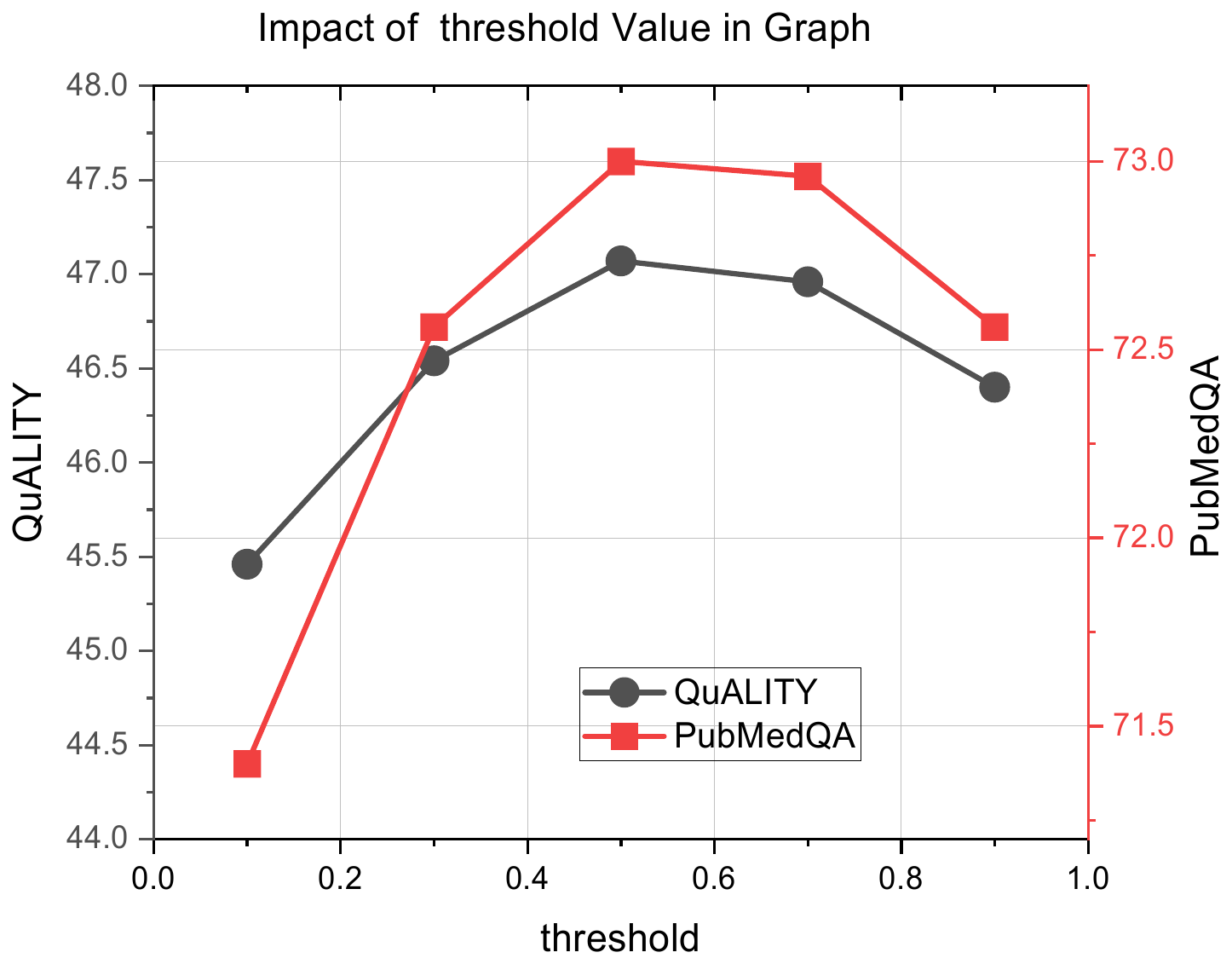}
    \caption{Impact of $\tau$ Value in Graph. }
    \label{fig:threshold}
\end{figure}

As the value of $ \tau $ increases, the number of edges decreases. As illustrated in Figure~\ref{fig:threshold}, performance reaches its maximum around a threshold of 0.5. Both metrics exhibit an upward trend as the threshold increases from 0.0 to approximately 0.5, peaking at this threshold before declining as the threshold exceeds 0.5. These observations indicate that both QuALITY and PubMedQA attain optimal values at a threshold of around 0.5. Performance for both metrics declines when the threshold is either too low or too high.  The fluctuation induced by $\tau$ is approximately 1 point, which does not significantly impact the robustness of our method.
\section{Contributions Analysis of Layers}
\label{sa_topk}
Statistical analysis of the top-k document distribution across different graph layers highlights the respective contributions of both low-level and high-level documents to the final output generation.
We retrieve 10 small documents (Top-k = 10) on a graph architecture with only 2 layers (n-layers =  2) for the test sets (LLM = llama3-8b).
\begin{table}[h]
\resizebox{\linewidth}{!}{
\begin{tabular}{l|llll}
\hline
PubMedQA             & 0-2   & 3-5   & 6-8   & 9-10  \\ \hline
layer 1 (low-level)  & 0     & 0.032 & 0.73  & 0.238 \\
layer 2 (high-level) & 0.592 & 0.394 & 0.014 & 0.0   \\ \hline
\end{tabular}
}
\caption{Contributions Analysis of the distribution of top-k documents across different graph layers on PubMedQA.}
\label{tab:ca1}
\end{table}
\begin{table}[h]
\resizebox{\linewidth}{!}{
\begin{tabular}{l|llll}
\hline
QuALITY              & 0-2   & 3-5    & 6-8   & 9-10  \\ \hline
layer 1 (low-level)  & 0.013 &	0.170 &	0.633 &	0.184 \\
layer 2 (high-level) & 0.415 &	0.505 &	0.079 &	0.001  \\ \hline
\end{tabular}
}
\caption{Contributions Analysis of the distribution of top-k documents across different graph layers on QuALITY.}
\label{tab:ca2}
\end{table}
Tables~\ref{tab:ca1} and~\ref{tab:ca2} show the proportion of test set samples with 10 retrieved relevant documents distributed across two layers of the graph network. For example, 0.592 means that out of the 10 retrieved documents, 0-2 of them fall on the second layer, accounting for 59.2\
From~\ref{tab:ca1} and~\ref{tab:ca2}, we can observe that the retrieved relevant documents may appear at any level. Some samples obtain more documents on the second layer than on the first layer after ranking, while others do the opposite. This indicates both low- and high-level information is important, which aligns with our motivation. Besides, the retrieved documents appear more in the first layer in general. This is mainly because the nodes in the second layer are usually fewer than those in the first layer.
\section{The Importance of Ranking}
\label{iranking}
\begin{table}[h]
\resizebox{\linewidth}{!}{
\begin{tabular}{lll}
\hline
                                               & \textbf{PubMedQA} & \textbf{QuALITY} \\ \hline
\multicolumn{1}{l|}{only llm w/o graph}        & 49.60             & 32.10            \\
\multicolumn{1}{l|}{only layer 1 (low-level)}  & 68.80             & 41.73            \\
\multicolumn{1}{l|}{only layer 2 (high-level)} & 62.60             & 40.23            \\
\multicolumn{1}{l|}{both layers}               & 73.00             & 47.05            \\ \hline
\end{tabular}
}
\caption{Generating outputs using different levels of layers.}
\label{tab:ranking}
\end{table}
From table~\ref{tab:ranking}, we can see that compared to not using a graph structure (only llm w/o graph), using only one layer of nodes (only layer 1 or layer 2) produces better results. Additionally, using only layer 1 (low-level) produces relatively better performance than using only layer 2 (high-level). Of course, using both low- and high-level information simultaneously (both layers) will achieve significant performance improvements. This indicates that: 1) It is necessary to design comprehensive graph networks and retrieval mechanisms to model external information, which is consistent with the motivation of this paper; 2) The amount of low-level information is greater, which is consistent with the distribution we calculated above. 3) Both low-level and high-level information have their own advantages, and utilizing them can achieve optimal performance. This motivates the design of our approach.
\section{Details of Building Graph Algorithm}
In the practical implementation, to save computational costs, we also select the top k edges by similarity measurement of the current node that connects to other nodes. The detailed process can refer to Algorithm~\ref{build-app}.
\begin{algorithm}[h]
\caption{Build Graph Algorithm}
\label{build-app}
\begin{algorithmic}[1]
\Function{Build Graph}{
$text$ \Comment{Document}, 
$large, small$,  \Comment{Output length}
$k$,  \Comment{Top K to build graph}
$n\_layers$, \Comment{Depth $ L $}
$\tau$ \Comment{Threshold}
}
\State $D_{large} \gets \text{TextSplit}(text,large)$
\State $S_D \gets \text{LLM.summay}(D_{large},small)$
\State $S_D'\gets \text{TextSplit}(D_{large},small)$
\State $D \gets S_D \cup S_D'$
\State $layers \gets []$
\While{$n\_layers >0$}
\State $e_D \gets \textsc{Embed}(D)$
\State $M \gets \text{sim}(e_D,e_D^\text{T})$
\State $sims \gets \text{max}_k(M, axis \gets 0)$
\State $\mathcal{V} \gets \text{arg~max}_k(M, axis \gets 0)$
\State $\mathcal{E} \gets []$
\For{$u \gets 0$ to $\text{len}(\mathcal{V})$}
    \For{$v ~~\text{in} ~~ \mathcal{V}[u]$}
        \State $w \gets sims[u][v]$
        \If {$w \ge \tau$}
        \State $\mathcal{E}.append((u, v, w))$
        \EndIf
    \EndFor
\EndFor
\State $\mathcal{G} \gets (\mathcal{V},\mathcal{E})$
\State$layers.\text{append}(\mathcal{G})$
\State $n\_layers \gets n\_layers-1$
\Comment{Prepare for the next layer}
\If{$n\_layers >0$}
    \State $C \gets \text{CommDetect}(\mathcal{G})$ 
    \State $D \gets \text{LLM.summay}(C,small)$
\EndIf
\EndWhile
\State \Return{layers} 
\EndFunction
\end{algorithmic}
\end{algorithm}
\onecolumn
\section{Case Study}
\subsection{A case of PubMedQA question answering} \label{case_study2}
\begin{casestudyblock}{Question:}
\sethlcolor{pink}
Treadmill testing of children who have \hl{spina bifida} and are ambulatory: does peak oxygen uptake reflect maximum oxygen uptake? 
\end{casestudyblock}
\begin{casestudyblock}{Our GraphMPA:}
\textbf{Context:}['The aim of this study was to determine whether the Vo(2)peak measured during an incremental treadmill test is a true reflection of the maximum oxygen uptake ((.)Vo(2)max) in children who have \hl{spina bifida} and are ambulatory.',
 'Earlier studies have demonstrated low peak oxygen uptake in children with \hl{spina bifida}. Low peak heart rate and low peak respiratory exchange ratio in these studies raised questions regarding the true maximal character of peak values obtained with treadmill testing...',
 'Twenty children who had \hl{spina bifida} and were ambulatory participated. ...',
 'The results of this study suggest that the peak measured during an incremental treadmill test is a 
\sethlcolor{yellow}
 \hl{valid indicator}
\sethlcolor{pink}
 of max in children who have \hl{spina bifida} and are ambulatory.',
 "The study was conducted in a pediatric emergency department (PED) in an urban hospital. ..",
]
\textbf{Reason:} The study aimed to determine if peak oxygen uptake during treadmill testing in 
children with \hl{spina bifida} accurately reflects their maximum oxygen uptake, considering previous studies raised 
questions about the true maximal character of peak oxygen uptake values obtained with treadmill testing in these 
children.
\textbf{Answer:} \textbf{Yes} {\color{green}\checkmark} 
\end{casestudyblock}
\begin{casestudyblock}{RAPTOR:}
    \textbf{Context:} Here's a summary of the study:  **Study Background and Objective:** The study aimed to determine whether the peak oxygen uptake (VO2peak) measured during an incremental treadmill test is a true reflection of the maximum oxygen uptake (VO2max) in children with \hl{spina bifida} who are ambulatory.  **Methodology:** A cross-sectional study was conducted with 20 children with \hl{spina bifida} who were ambulatory. The VO2peak was measured during a graded treadmill exercise test... 
\textbf{Answer:} \textbf{maybe} {\color{red}\texttimes}
\end{casestudyblock}
\begin{casestudyblock}{LightRAG:}
\textbf{Answer:} \textbf{Maybe} {\color{red}\texttimes}
The inverse equity hypothesis suggests that new health interventions are initially 
primarily accessed by the rich, but that inequalities narrow with diffusion to the poor. However, the applicability
of this hypothesis to coverage of cancer screening in middle-income countries is uncertain.
According to the 
data, a study examined the relationship between overall coverage and economic inequalities in coverage of cancer 
screening in four middle-income countries. The results showed that economic inequalities in screening participation
were substantially lower or non-existent in countries with higher aggregate coverage, for both breast cancer 
screening and cervical cancer screening.
...
\end{casestudyblock}
Above are case study comparisons of answering models. The responses from three different models: Ours, RAPTOR, and LightRAG to the question from PubmedQA~\cite{jin2019pubmedqadatasetbiomedicalresearch}. The keywords are highlighted. Our model (highlighted in green) correctly answers "Yes",  In contrast, both RAPTOR and LightRAG provide uncertain and incorrect responses ("maybe"), demonstrating the difference in performance and confidence between the models.
\newpage
\subsection{A case of graph constructing} \label{case_study3}
\begin{figure}[h!]
    \centering
    \begin{subfigure}[b]{0.45\textwidth}
        \includegraphics[width=\textwidth]{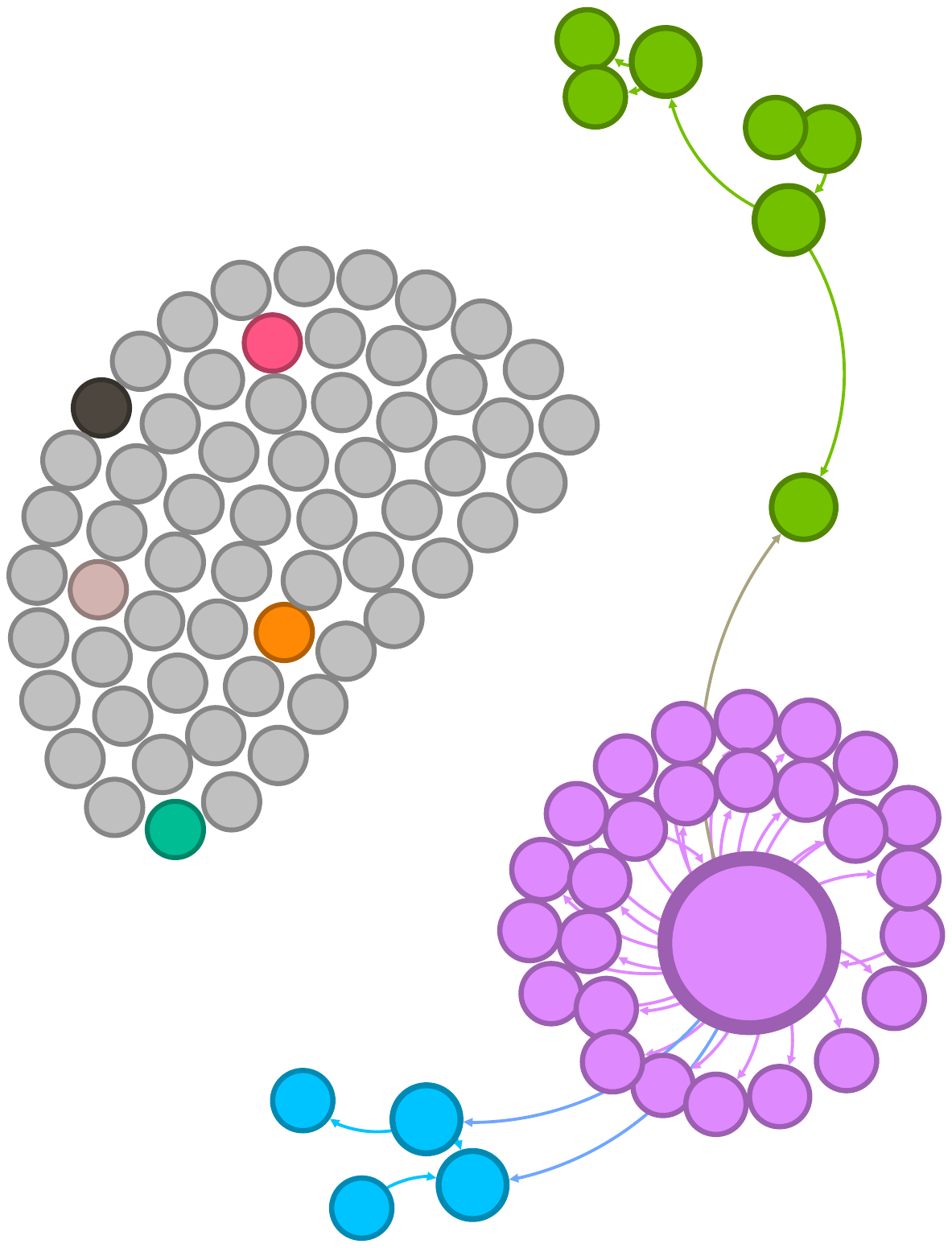}
    \caption{LightRAG}
    \end{subfigure}
    \hfill
    \begin{subfigure}[b]{0.45\textwidth}
        \includegraphics[width=\textwidth]{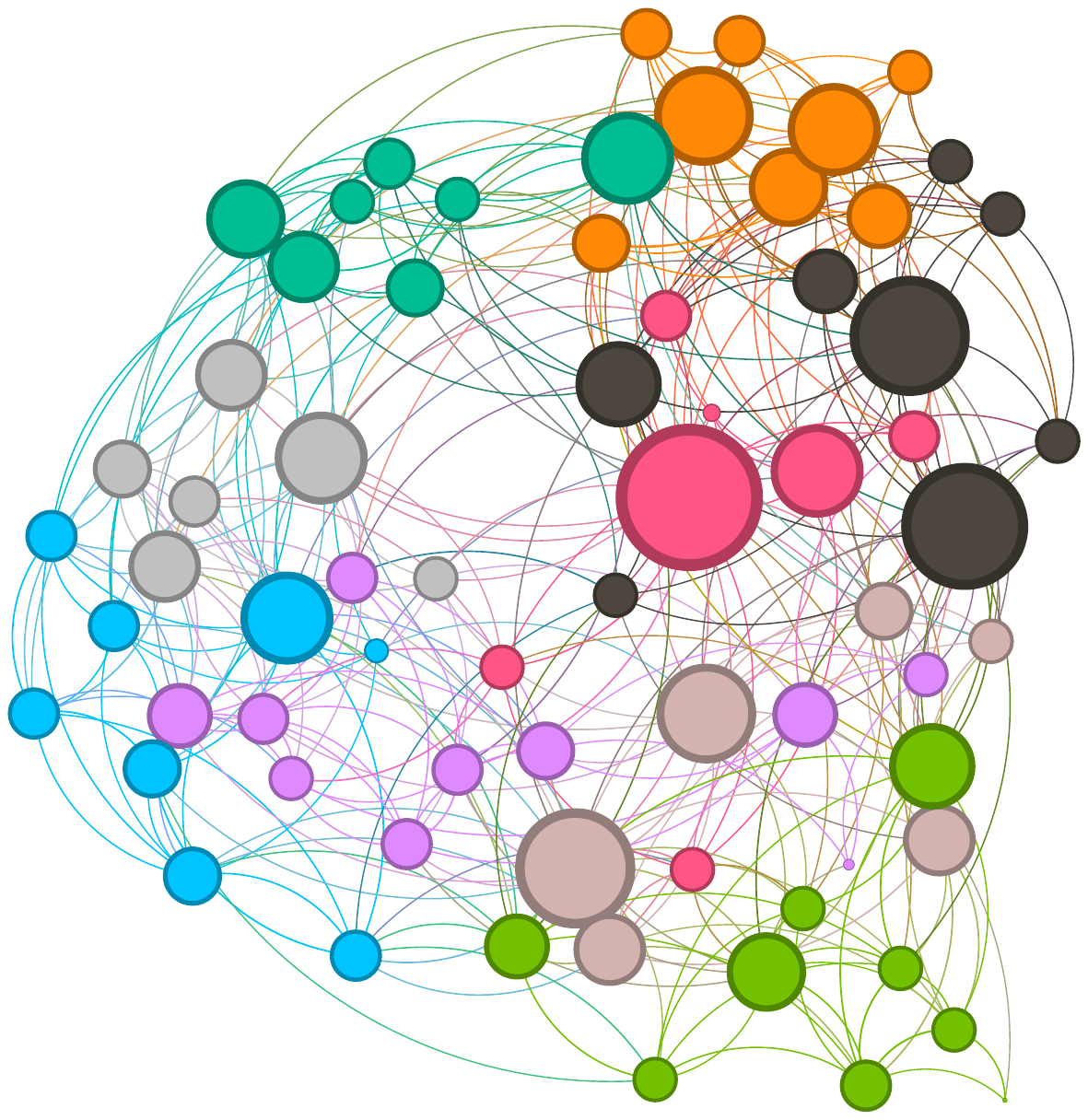}
    \caption{Our GraphMPA}
    \end{subfigure}
    \label{fig:graph_light_our}
    \caption{An example of QuALITY, LightRAG extracts nodes and edges from documents compared to our GraphMPA build graph by node similarity. GraphMPA generates nodes with rich edges while LightRAG extracts many isolated nodes.}
\end{figure}
We randomly select a paper from QuALITY to construct a graph using LightRAG and our GraphMPA. Notably, LightRAG extracts many isolated nodes, while GraphMPA generates nodes with rich edges. Then we use two methods to choose the option corresponding to the question (below)
based on the graph. Due to the limited ability of LightRAG to accurately capture node relations, it struggles with relation leakage to generate the correct answer 
. In contrast, our GraphMPA effectively captures and generates with greater accuracy.
The key reason is highlighted.
\begin{tcolorbox}
\label{case2_question}
\textbf{Question:}
Why is Si retirement so significant to the Space Exploration Team? 
\textbf{Options: }
1. There aren’t enough working people in the world. They won’t be able to find a replacement.
2. As one of two remaining spacemen, it would likely mean the defunding and shut down of the Space Exploration Team.
3. Training new spacemen is costly and time consuming. They won’t have anyone else ready after him.
4. His retirement may inspire others to stop working as well, which would be hugely detrimental as most people don't feel the drive to work as is.  
\end{tcolorbox}
\begin{tcolorbox}[title={LightRAG}]
\textbf{Process:}
high\_level\_keywords: ["Significance of retirement", "Space Exploration Team", "Workforce impact", "Team dynamics"]
low\_level\_keywords: ["Si retirement", "Spacemen", "Defunding", "Shut down", "Training costs", "Replacement", "Workforce shortage", "Motivation"]
Local query uses 60 entites, 38 relations, 3 text units.
Global query uses 38 entites, 38 relations, 2 text units.
\tcblower
\textbf{Output:}
**The Significance of Si's Retirement to the Space Exploration Team**
Si's retirement is a significant event for the Space Exploration Team, and it's not just because he's a skilled space pilot. \hl{The team is facing a critical shortage of working people, and Si's retirement would exacerbate this issue.} Here are some reasons why his retirement is so crucial to the team:
\subsection{Lack of Replacement}
The team is struggling to find a replacement for Si, and his retirement would only worsen the situation. With a dwindling pool of skilled workers, the team would be left without a capable pilot to take on the challenges of space exploration.
\subsection{Defunding and Shutdown}
As one of the last two remaining spacemen, Si's retirement would likely lead to the defunding and shutdown of the Space Exploration Team. The team's existence relies on the availability of skilled pilots like Si, and without him, the program would become unsustainable.
\subsection{Costly and Time-Consuming Training}
Training new spacemen is a costly and time-consuming process. The team has invested significant resources in Si's training, and his retirement would mean that those resources would be wasted. Moreover, it would take years to train a new pilot to Si's level of expertise, and even then, there's no guarantee that they would be as skilled or dedicated as Si.
\subsection{Inspiring Others to Stop Working}
Si's retirement may also have a ripple effect, inspiring others to stop working as well. In a society where most people don't feel the drive to work, Si's decision to retire could be seen as a validation of their own apathy. This could lead to a further decline in productivity and a worsening of the already dire situation.
The Space Exploration Team is facing a perfect storm of challenges, and Si's retirement would be the final blow. His decision to retire is not just a personal choice; it has far-reaching consequences for the team and the future of space exploration.
**The Consequences of Si's Retirement**
Si's retirement would have significant consequences for the Space Exploration Team, including:
\begin{itemize}
\item **Loss of expertise**: Si's retirement would mean the loss of his expertise and experience, which would be difficult to replace.
\item **Defunding and shutdown**: The team's existence would be threatened, and it may be forced to shut down due to the lack of a skilled pilot.
\item **Inspiring others to stop working**: Si's retirement could inspire others to stop working, exacerbating the already dire situation.
\item **Impact on the economy**: The loss of a skilled pilot like Si would have a ripple effect on the economy, leading to a decline in productivity and a worsening of the already dire situation.
\end{itemize}
The Space Exploration Team is facing a critical moment, and Si's retirement is a significant factor in this crisis. The team needs to find a way to convince Si to stay on or find a suitable replacement to avoid the consequences of his retirement.",
\textbf{Extract Answer}: 2 {\color{red}\texttimes}
\textbf{Global Truth}: 3
\label{case2_lightrag}
\end{tcolorbox}
\begin{tcolorbox}[title={Our GraphMPA}]
\textbf{Context:}
["Si Pond was a space pilot, and he was retiring.  He was at a bar with a woman who was clearly still in shock from meeting him.  She had cried when he was given an antique watch, and was now avoiding his face.  She mentioned a speech by Doctor Girard-Perregaux, who had praised Si for his accomplishments as a space pilot.  Si downplayed his achievements, saying that two of his runs were only to the Moon.  The woman was not amused by Si's attitude, and he offered to buy her another drink.  The woman was clearly still in awe of Si, and was struggling to come to terms with the fact that he was retiring.  Si, on the other hand, seemed to be taking his retirement in stride, and was more concerned with the politics of the Space Exploration department than with his own accomplishments.  He mentioned that the department was in danger of being dropped by the Appropriations Committee, and that his retirement was part of a larger scheme to pressure him into taking on more trips.  The woman was not impressed by Si's cynicism, and the conversation ended with him offering to buy her another drink.  Key details:  * Si Pond is a space pilot who is retiring.", 
' There you stood, so fine and straight in  your space-pilot uniform, the veteran of six exploration runs to the  planets "  "Well," Si said modestly, "two of my runs were only to the Moon "  "  and he said all those things about man\'s conquest of space  And  the dream of the stars which man has held so long  And then the fact  that \hl{ you were the last of the space pilots  The last man in the whole}', 
" They also had a banquet for him, complete with speeches by such  bigwigs of the Department of Space Exploration as Academician Lofting  Gubelin and Doctor Hans Girard-Perregaux  There was also somebody  from the government who spoke, but he was one of those who were  pseudo-elected and didn't know much about the field of space travel  nor the significance of Seymour Pond's retirement  Si didn't bother to  remember his name", 
' Si said, "Look, how about another drink "  Natalie Paskov said, "Oh, I\'d love to have a drink with you, Mr " "Si," Si said  He motioned to the bartender with a circular twist of  the hand indicating their need for two more of the same  "How come you  know so much about it  You don\'t meet many people who are interested  in space any more', 
" They hadn't figured he had enough shares of Basic to see him  through decently  Well, possibly he didn't, given their standards  But Space Pilot Seymour Pond didn't have their standards  He'd had  plenty of time to think it over  It was better to retire on a limited  crediting, on a confoundedly limited crediting, than to take the two or  three more trips in hopes of attaining a higher standard"]
\tcblower
\textbf{Reason: }Si's retirement marks the end of an era in space exploration, and he is the last of a breed of space pilots. 
\textbf{Answer: }The significance of Si's retirement is that it marks the end of an era in space exploration and the last of a breed of space pilots.
\textbf{Extract Answer}: 3 {\color{green}\checkmark} 
\textbf{Global Truth}: 3
\label{case2_ours}
\end{tcolorbox}
\section{Details of Train Data Example}
\label{train_data}
\begin{tcolorbox}
\textbf{Question $q$}
Does histologic chorioamnionitis correspond to clinical chorioamnionitis?
\noindent
\textbf{Context $C$}
[
'To evaluate the degree to which histologic chorioamnionitis, a frequent finding in placentas submitted for histopathologic evaluation, correlates with clinical indicators of infection in the mother.',
'A retrospective review was performed on 52 cases with a histologic diagnosis of acute chorioamnionitis from 2,051 deliveries at University Hospital, Newark, from January 2003 to July 2003. Third-trimester placentas without histologic chorioamnionitis (n = 52) served as controls. Cases and controls were selected sequentially. Maternal medical records were reviewed for indicators of maternal infection.',
'Histologic chorioamnionitis was significantly associated with the usage of antibiotics (p = 0.0095) and a higher mean white blood cell count (p = 0.018). The presence of 1 or more clinical indicators was significantly associated with the presence of histologic chorioamnionitis (p = 0.019).'
]
\tcblower
\textbf{Chosen answer $y_w$}
\#\#\#Reason: Histologic chorioamnionitis is a reliable indicator of infection whether or not it is clinically apparent.
\#\#\#Answer: yes
\textbf{Rejected answer $y_l$}
yes
\end{tcolorbox}
\end{document}